%% file: main.tex
\documentclass{article}

\usepackage{PRIMEarxiv}
\usepackage[utf8]{inputenc}
\usepackage[T1]{fontenc}
\usepackage{booktabs}
\usepackage{amsfonts}
\usepackage{nicefrac}
\usepackage{microtype}
\usepackage{fancyhdr}
\usepackage{graphicx}
\usepackage{placeins}
\usepackage{natbib}
\usepackage{amsmath}
\usepackage{subcaption}
\usepackage{rotating}
\usepackage[table]{xcolor}
\usepackage{algorithm}
\usepackage{algpseudocode}
\usepackage{dsfont}
\definecolor{darkblue}{RGB}{0,0,128}
\usepackage{hyperref}
\hypersetup{
    colorlinks=true,
    linkcolor=darkblue,
    citecolor=darkblue,
    urlcolor=darkblue
}
\title{Probing the Probes: Methods and Metrics for Concept Alignment}

\author{
  Jacob Lysnæs-Larsen, Marte Eggen, Inga Strümke \\
  Department of Computer Science \\
  NTNU - Norwegian University of Science and Technology \\
  Trondheim, Norway\\
  \texttt{\{jacob.lysnas-larsen, marte.eggen, inga.strumke\}@ntnu.no} \\
}

\begin{document}

\FloatBarrier

\maketitle

\begin{abstract}
    In explainable AI, Concept Activation Vectors (CAVs) are typically obtained by training linear classifier probes to detect human-understandable concepts as directions in the activation space of deep neural networks. It is widely assumed that a high probe accuracy indicates a CAV faithfully representing its target concept. However, we show that the probe's classification accuracy alone is an unreliable measure of concept alignment, i.e., the degree to which a CAV captures the intended concept. In fact, we argue that probes are more likely to capture spurious correlations than they are to represent only the intended concept. As part of our analysis, we demonstrate that deliberately misaligned probes constructed to exploit spurious correlations, achieve an accuracy close to that of standard probes. To address this severe problem, we introduce a novel concept localization method based on spatial linear attribution, and provide a comprehensive comparison of it to existing feature visualization techniques for detecting and mitigating concept misalignment. We further propose three classes of metrics for quantitatively assessing concept alignment: hard accuracy, segmentation scores, and augmentation robustness. Our analysis shows that probes with translation invariance and spatial alignment consistently increase concept alignment. These findings highlight the need for alignment-based evaluation metrics rather than probe accuracy, and the importance of tailoring probes to both the model architecture and the nature of the target concept.
\end{abstract}


\FloatBarrier
\section{Introduction}
Concept-based explanation methods have over the past years established themselves as a promising approach to understand the inner workings of deep neural networks. These methods aim to translate non-interpretable neural activations into human-understandable concepts, such as \texttt{stripes}, \texttt{woman}, or \texttt{building}, that are not explicitly present in the input data. A common technique for detecting concepts is by training linear classifier probes to learn activation patterns that separate activations by concept. Concept detection has proven useful for a wide range of downstream tasks, including quantifying concept importance \citep{Kim2018InterpretabilityTCAV}, creating interpretable concept bottleneck models \citep{Koh2020ConceptModels}, detecting adversarial attacks \citep{Li2025InterpretableVector}, mitigating model bias \citep{Wu2023DiscoverCorrelation, Joo2024DebiasedRegularization}, and generating counterfactual explanations \citep{Abid2022MeaningfullyExplanations}.

For concept-based explanations to be meaningful, probes must accurately learn the intended concepts, a property we refer to as \emph{concept alignment}. We identify two primary failure modes. In the simplest case, the activations do not contain linearly available information about the target concept. This may occur because the neural network has not internalized the concept at all, or because the concept is encoded non-linearly. Either way, the probe will fail to separate activations by concept, and it will be easily detectable as it leads to probes with poor classification accuracy. In the more challenging case, the probe learns features correlated with the concept rather than the concept itself, or a combination of both. Given spurious correlations that are highly predictive of the target concept, it is possible for the probe to fail to learn the target concept yet still obtain high classification accuracy. In this work, we focus specifically on the underexplored failure mode involving spurious correlations.

In our study, we investigate what probes actually learn, and use for demonstration purposes a widely used deep Convolutional Neural Network (CNN). We focus on linear probes, specifically those producing Concept Activation Vectors (CAVs), as introduced in \citet{Kim2018InterpretabilityTCAV}, where it is assumed that concepts are represented as directions in activation space, i.e., linear combinations of neurons. We show that with a simple procedure, probes can be trained without using examples of the target concept in the training data and still achieve an average classification accuracy of $74\%$ compared to $81\%$ for standard probes (averaged over $148$ concepts). Consequently, we argue that the probe's classification accuracy is a potentially misleading metric, and inadequate for measuring concept alignment, despite being commonly used in the literature to assess the `quality' of the associated CAV, see for instance \citet{Kim2018InterpretabilityTCAV, Arendsen2020ConceptScenicness, Abid2022MeaningfullyExplanations, Crabbe2022ConceptExplanations, Bai2023CONCEPTASSUMPTION, Gao2024GeneratingSensitivity, Li2025InterpretableVector, Schmalwasser2025FastCAV:Networks}. We further evaluate alternative probing methods for obtaining CAVs, including established and novel variants. The probing methods are analyzed both qualitatively and quantitatively. First, we use feature visualization techniques to better understand what learned CAVs represent, thereby demonstrating different inductive biases for each probing method. We then introduce novel metrics to assess concept alignment, presenting a comparative summary in tabular form alongside an analysis of how alignment scales with the size of the concept training data. In summary, our main contributions are:

\begin{itemize}
    \item An analysis demonstrating that existing probing methods are prone to produce CAVs that are misaligned with their target concepts. We further identify the dominating cause of this misalignment to be spurious correlations.

    \item Upon demonstrating that probe classification accuracy alone is unreliable to assess concept alignment, we propose novel metrics and visualization methods.

    \item Using our proposed visualization methods and metrics, we show that probes that incorporate translation invariance and spatial alignment generally result in more aligned CAVs.
\end{itemize}

The remainder of this paper is organized as follows: Section~\ref{sec:related_work} summarizes relevant prior work. Section~\ref{sec:concept_misalignment} demonstrates the prevalence of concept misalignment, followed by Section~\ref{sec:obtaining_and_visualizing}, which presents methods for obtaining and interpreting CAVs. In Section~\ref{sec:quantifying_alignment}, concept alignment is quantified using metrics for robustness, accuracy, and segmentation. Section~\ref{sec:discussion} discusses implications of the results, limitations, and future work, and Section~\ref{sec:conclusion} concludes the paper. 

\section{Related work}\label{sec:related_work}
While the encoding of concepts in neural network activations is not fully understood, work such as \citet{Zhou2015ObjectCNNs, Bau2017NetworkRepresentations, goh2021multimodal} show that individual neuron activations often correlate with human-understandable concepts. This is often referred to as the activation space having a \emph{privileged basis}, namely that concepts are inherently more likely to be encoded in the basis directions, i.e., individual neurons, rather than arbitrary directions in activation space. However, other studies find no difference in interpretability between the basis directions and random directions \citep{Szegedy2013IntriguingNetworks}, challenging the idea of neurons being privileged. Later work by \citet{elhage2022toy} argues that whether neurons form a privileged basis depends on the network architecture. Furthermore, \citet{olah2020zoom} hypothesize that neurons encode concepts in \emph{superposition}. Their superposition hypothesis aims to explain how neural networks are able to represent more concepts than it has neurons available. Although an $n$-dimensional space supports at most $n$ orthogonal vectors, it can contain exponentially many approximately orthogonal vectors \citep{elhage2022toy}. We work under the assumption that concepts are encoded as arbitrary directions, and aim to learn them by linear probing, yet we acknowledge ongoing work on non-linear encodings, e.g., \citet{Zaeem2021CauseNetworks, Crabbe2022ConceptExplanations, Bai2023CONCEPTASSUMPTION, Engels2025NotLinear}.

\citet{Alain2017UNDERSTANDINGPROBES} propose linear classifier probes for monitoring class information throughout layers, and \citet{Kim2018InterpretabilityTCAV} generalize the approach by replacing output classes with human-understandable concepts, provided datasets of the target concept and of negative examples. Classifier probes are typically trained using logistic regression or linear Support Vector Machines (SVMs), and is arguably the most widely used supervised approach for obtaining CAVs, as seen in works such as \citet{Kim2018InterpretabilityTCAV, Lucieri2020ExplainingMaps, Abid2022MeaningfullyExplanations, McGrath2022AcquisitionAlphaZero, Gupta2023ConceptImprovement, Joo2024DebiasedRegularization, Dreyer2024FromSpace, Li2025InterpretableVector}. In a similar line of work to classifier probes, \citet{Bau2017NetworkRepresentations} and \citet{Fong2018Net2Vec:Networks} learn CAVs by matching feature maps with segmentation masks of the target concepts. Alternative approaches include computing the difference between mean activations of samples containing the concept and those without it \citep{Brocki2019ConceptModels, DeSantis2024Visual-TCAV:Classification, Pahde2025NavigatingDivergence, Schmalwasser2025FastCAV:Networks}. Notably, \citet{Pahde2025NavigatingDivergence} formalize this method as Pattern-CAV, and demonstrate that it is more robust to noise than linear classifier probes. Recently, a growing body of work has also explored how Vision Language Models (VLMs) can be used to obtain CAVs \citep{Moayeri2023Text2Concept:Text, Moayeri2023Text-To-ConceptAlignment, Huang2024LG-CAV:Guidance, Nicolson2024TextCAVs:Text}. Although these CAVs are derived from VLM activation spaces, \citet{Moayeri2023Text-To-ConceptAlignment} show that CAVs can often be linearly mapped between models.

Classifier probes are generally known to be vulnerable to spurious correlations, see e.g., \citet{Belinkov2022ProbingAdvances} for a detailed discussion in the context of natural language processing, and the challenges described by \citet{McGrath2022AcquisitionAlphaZero} in the context of chess. However, to our knowledge, no prior work has extensively and empirically investigated methods for detecting and mitigating spurious correlations in both classifier and alternative probing methods. It is therefore unclear to what extent concept misalignment occurs, when it happens, and how to detect, evaluate and prevent it. Instead, much of the explainable AI literature focuses on developing downstream applications that implicitly assume well-aligned CAVs. Research on concept-based methods often addresses issues related to concept misalignment but rarely studies it directly. Related work in concept alignment includes, most notably, dataset selection \citep{Ramaswamy2023OverlookedCapability}, noise robustness \citep{Pahde2025NavigatingDivergence}, information leakage in concept learning models \citep{Mahinpei2021PromisesModels, Margeloiu2021DoIntended}, and robustness to random seeds \citep{Soni2020AdversarialNetworks}.

Significant efforts have been made to develop feature visualization techniques to understand what directions in activation space represent, but these methods are rarely used to interpret CAVs obtained by supervised probing methods. When supervised probes are used to learn a target concept, users are susceptible to confirmation bias and may overconfidently assume that the probe captures the target concept. As for probing with unsupervised learning, see for instance \citet{Ghorbani2019TowardsExplanations, Yeh2020OnNetworks, Zhang2021InvertibleVectors, Gorton2024TheVision}, the learned CAVs must be interpreted post-hoc, as there is no user-defined target concept. As a result, CAVs discovered through unsupervised methods are often scrutinized, whereas those obtained via supervised methods are rarely questioned. We build upon the work of unsupervised concept discovery and use feature visualization techniques to interpret CAVs obtained by supervised probes.

The arguably most common approach to interpreting CAVs is to visualize images that result in highly similar activations to it \citep{Kim2018InterpretabilityTCAV, Arendsen2020ConceptScenicness, Zhang2021InvertibleVectors, Mikriukov2024UnveilingOfCNNs, Hossain2024EnhancingModels, Akpudo2024Coherentice:Fidelity, Schmalwasser2025ExploitingConcepts}. Importantly, \citet{Fong2018Net2Vec:Networks} argue for visualizing a variety of quantiles to avoid overrepresenting extreme inputs. Instead of restricting visualizations to existing images, many methods have been developed to generate synthetic images that result in activations highly similar to CAVs \citep{olah2017feature}. However, \citet{Zimmermann2021HowActivations} found that such synthetic images offer little explanatory advantage over highly activating natural images. Additionally, \citet{Geirhos2023DontVisualizations} find that CNNs process synthetic images distinctly from natural ones, questioning their usefulness for interpretability. There is also an increasing popularity of using saliency methods via gradients to highlight concept-relevant features, including \citet{Brocki2019ConceptModels, Lucieri2020ExplainingMaps, Zhang2021InvertibleVectors, Margeloiu2021DoIntended, Akpudo2024Coherentice:Fidelity, Dreyer2024FromSpace, DeSantis2024Visual-TCAV:Classification, Aysel2025Concept-BasedBenchmarks}. Some quantitative metrics have also been developed to assess the completeness of a set of concepts \citep{Yeh2020OnNetworks, EspinosaZarlenga2023TowardsEvaluation}.

\FloatBarrier
\section{Demonstrating concept misalignment}\label{sec:concept_misalignment}
\begin{figure}
        \centering
        \includegraphics[width=1\linewidth]{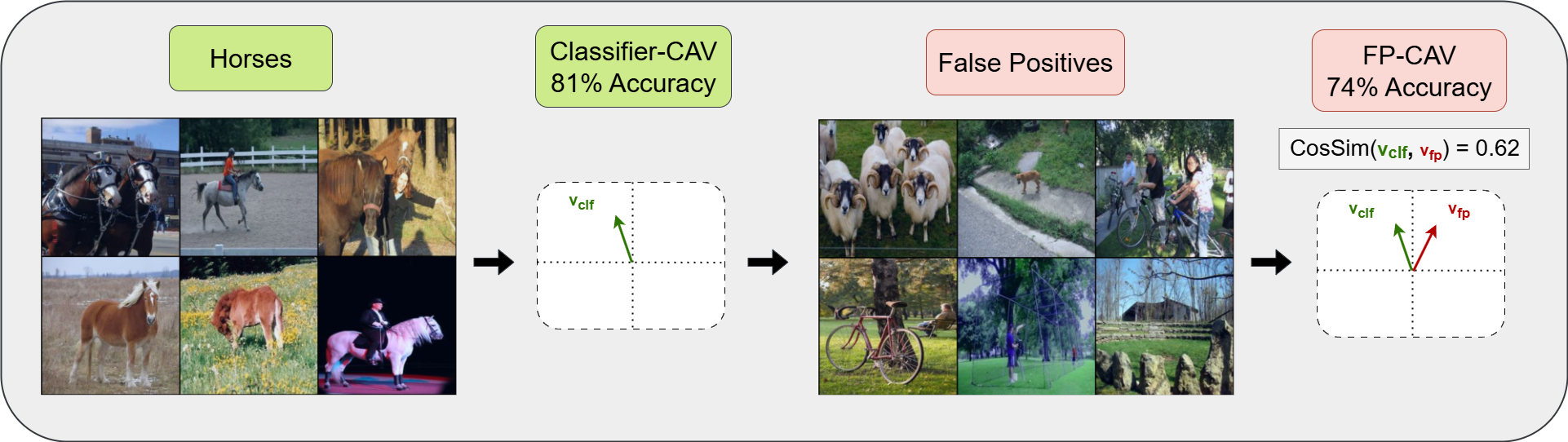}
        \caption{Illustration of the procedure used to create FP-CAVs, which demonstrates the prevalence of concept misalignment. We first train a standard classifier probe on the concept \texttt{horse} and collect a set of false positives. We then train a second classifier on those false positives, using incorrect target labels. When both classifiers are tested on the same held-out dataset, they achieve similar classification accuracy. Additionally, the two corresponding CAVs have a high cosine similarity. This shows that classifier probes learn significant amounts of spurious correlations and that accuracy is unreliable for measuring concept alignment.}
        \label{fig:fp_summary}
\end{figure}

\subsection{Model and data}\label{sec:model_data}
In this work, we identify concept directions in the activation space of ResNet50 \citep{He2016DeepRecognition} pretrained on ImageNet \citep{Deng2009ImageNet:Database}. As object-level information is often concentrated in the later parts of CNNs \citep{Kim2018InterpretabilityTCAV, cammarata2020thread:}, we extract post-ReLU activations from the first bottleneck block in the final layer. Concept images and corresponding segmentation masks are taken from the Broden dataset \citep{Bau2017NetworkRepresentations}. Specifically, we select concepts in the category of objects that appear in at least $150$ images and occupy at least $1\%$ of the image area. This yields $148$ concepts in total, including concepts like \texttt{building}, \texttt{dog}, and \texttt{car}. For each concept, we define positive samples as images where the concept is present, and negative samples as images where it is absent. In our experiments, we use $50$ positive and $50$ negative samples to train probes, and $100$ positives and $100$ negatives for evaluation, unless otherwise specified. We only use balanced datasets, where the random baseline corresponds to a classification accuracy of $0.5$.

\subsection{Binary linear classifier probes}\label{sec:linear_clf_probes}
We consider a neural network $f: \mathbb{R}^{d_0} \rightarrow \mathbb{R}^{d_L}$ that performs a mapping from the input layer $l=0$ to the output layer $l=L$. Within the neural network, we are interested in the concept directions in some target layer $l \in \{0, 1, \dots, L\}$ and denote a feature extractor $f_l: \mathbb{R}^{d_0} \rightarrow \mathbb{R}^{d_l}$ to collect the corresponding activations. To learn a CAV for a concept $c$ in layer $l$, denoted $\mathbf{v}_{l,c}$, we use a dataset $\mathcal{X}$ of images and extract the subsets $\mathcal{P}_c \subset \mathcal{X}$ and $\mathcal{N}_c \subset \mathcal{X}$, corresponding to positive and negative samples, respectively. We further collect the activations $Z_{l,c}^+ = \left\{ f_l(\mathbf{x}) \mid \mathbf{x} \in \mathcal{P}_c \right\}$ and $Z^-_{l,c} = \left\{ f_l(\mathbf{x}) \mid \mathbf{x} \in \mathcal{N}_c \right\}$, representing the presence and absence of the concept. To simplify notation and avoid clutter, we hereafter omit the explicit indices for the concept $c$ and target layer $l$ in CAVs $\mathbf{v}_{l,c}$, activations $\mathbf{z}_{l,c}$, and sets $Z^+_{l,c}$, $Z^-_{l,c}$.

Intuitively, linear classifier probes aim to learn weights $\mathbf{v}$ and a bias term $b$ such that a hyperplane separates activations in $Z^+$ and $Z^-$ by
\begin{equation}    
    \mathbf{z} \cdot \mathbf{v} + b > 0 \quad \forall \mathbf{z} \in Z^+ \quad \text{and} \quad \mathbf{z} \cdot \mathbf{v} + b < 0 \quad \forall \mathbf{z} \in Z^- \,.
\end{equation}
Here, $\mathbf{v}$ is the normal vector of the hyperplane, aligned with the concept direction. In practice, $\mathbf{v}$ is commonly learned using a linear SVM classifier or logistic regression, possibly with $L_1$ or $L_2$ regularization. Including regularization terms can be interpreted as imposing Bayesian prior distributions on the CAV. Specifically, $L_1$ corresponds to a Laplace distribution $v_j \sim \text{Laplace}(0, \beta)$ for all $j$, whereas $L_2$ corresponds to a normal distribution $\mathbf{v} \sim \mathcal{N}(\mathbf{0}, \sigma^2 I)$, with the parameters $\beta > 0$ and $\sigma^2 > 0$ scaling the regularization strength. Such priors can be useful in cases where there are few data samples, but the distributions of activations that represent concepts in neural networks are generally unknown. Thus, no regularization is used throughout this work. Interestingly, we observe that CAVs trained on pre-ReLU activations tend to follow a normal distribution, whereas post-ReLU CAVs closely follow a Laplace distribution, see Appendix~\ref{ap:clf_cavs_distributions} for examples. As we do not consider these CAVs to be ground truth concept directions, we do not use these observations to impose priors.

Due to the high-dimensional activation space and relatively low amount of input samples, i.e., $d_l \gg |\mathcal{P}_c| + |\mathcal{N}_c|$, we find empirically that all $148$ tested concepts are perfectly linearly separable using logistic regression, that is, every training sample is correctly classified. In cases of perfect linear separability and no regularization, the negative log-likelihood of the weights $\mathbf{v}$ has no finite minimum, causing the weights to diverge to infinity. Hence, interpreting the magnitude of $\mathbf{v}$ becomes challenging. Therefore, we focus solely on the direction $\frac{\mathbf{v}}{||\mathbf{v}||}$ throughout this paper. Additionally, since there is no finite minimum, optimizers may converge to different solutions due to their implicit biases, even though the loss function is convex. For homogeneous logistic regression trained on linearly separable data, optimized by stochastic gradient descent (SGD), $\frac{\mathbf{v}}{||\mathbf{v}||}$ converges towards the direction of the maximum-margin solution, similar to linear SVMs \citep{Soudry2018TheData}. In contrast, optimizers like Adam \citep{Kingma2015Adam:Optimization}, using adaptive learning rates, result in significantly different solutions when comparing distributions (examples are shown in Appendix~\ref{ap:clf_cavs_distributions}). For our experiments, we use binary logistic regression without regularization optimized with Scikit-learn \citep{scikit-learn}. Mathematically, this is achieved by minimizing the binary cross-entropy
\begin{equation}
    \mathcal{L}_{\text{clf}}(\mathbf{v}, b) = 
    -\frac{1}{|Z^+|} \sum_{\mathbf{z}^+ \in Z^+} \log \sigma(\mathbf{v} \cdot \mathbf{z}^+ + b) - 
    \frac{1}{|Z^-|} \sum_{\mathbf{z}^- \in Z^-} \log (1 - \sigma(\mathbf{v} \cdot \mathbf{z}^- + b)) \\,
    \label{eq:classifier_probe}
\end{equation}
where $\sigma$ denotes the sigmoid function.

\subsection{False positive CAVs}\label{sec:fp_cavs}
In this section, we demonstrate that Classifier-CAVs $\mathbf{v}_{\text{clf}}$, i.e., those obtained by linear classifier probes as outlined in Section~\ref{sec:linear_clf_probes}, often obtain high classification accuracy while heavily relying on spurious correlations. We consider a feature to be spuriously correlated with a concept if it is predictive of the concept but not causally necessary for it. To demonstrate the prevalence of spurious correlations, we train $\mathbf{v}_{\text{clf}}$ to separate activations in $Z^+$ and $Z^-$, using balanced datasets $|Z^+| = |Z^-| = N$. We then compare each $\mathbf{v}_{\text{clf}}$ to a deliberately misaligned False Positive CAV (FP-CAV) $\mathbf{v}_{\text{fp}}$, trained to detect features that are correlated but not causally necessary for the concept. To achieve this, we identify and classify negative samples until $|Z^-_{\text{fp}}| = 50$, then train $\mathbf{v}_{\text{fp}}$ using those activations. The full procedure is detailed in Algorithm~\ref{alg:false_pos_cav} and Figure \ref{fig:fp_summary} provides an illustration.

\begin{algorithm}
    \caption{Procedure for obtaining FP-CAVs}
    \label{alg:false_pos_cav}
    \begin{algorithmic}[1]
        \State $Z^{-}_{\text{fp}} \gets \emptyset$ \Comment{Set for false positives, initially empty}
        \State $\theta \gets$ logistic regression hyperparameters
        \State \textbf{Let} $N$ be the number of training samples
        \State \textbf{Let} $M$ be the number of testing samples
        \State \textbf{Let} $K$ be the number of possible hard negatives samples \Comment{$K>N$, Value depends on probing accuracy}
        \State \textbf{Input:} $Z^{+}_{\text{train}}, Z^{-}_{\text{train}} \in \mathbb{R}^{N \times C \times H \times W}$
        \State \textbf{Input:} $Z^{+}_{\text{test}}, Z^{-}_{\text{test}} \in \mathbb{R}^{M \times C \times H \times W}$
        \State \textbf{Input:} $Z^{-}_{\text{buffer}} \in \mathbb{R}^{K \times C \times H \times W}$ 

        \Statex
        \Function{CreateClassifierCAV}{$Z^{+}, Z^{-}$}
            \State $\mathbf{v}, b \gets \text{LogisticRegression}(Z^{+}, Z^{-}, \theta)$
            \State $b \gets \frac{\mathbf{b}}{||\mathbf{v}||}$
            \State $\mathbf{v} \gets \frac{\mathbf{v}}{||\mathbf{v}||}$
            \State $\text{acc} \gets \text{ClassifierAccuracy}(\mathbf{v}, b, Z^{+}_{\text{test}}, Z^{-}_{\text{test}})$
            \State \Return $\mathbf{v}, b, \text{acc}$
        \EndFunction
        \Statex
        
        \State $\mathbf{v}_{\text{clf}}, b_{\text{clf}}, \text{acc}_{\text{clf}} \gets$ \Call{CreateClassifierCAV}{$Z^{+}_{\text{train}}, Z^{-}_{\text{train}}$}
        
        \State $i \gets 0$
        \While{$|Z^{-}_{\text{fp}}| < N$}
            \If{$\mathbf{v}_{\text{clf}} \cdot Z^{-}_{\text{buffer}}[i] + b_{\text{clf}} > 0$} \Comment{Check if sample is falsely classified as positive}
                \State Append $Z^{-}_{\text{buffer}}[i]$ to $Z^{-}_{\text{fp}}$
            \EndIf
            \State $i \gets i + 1$
        \EndWhile
        
        \Statex
        \State $\mathbf{v}_{\text{fp}}, b_{\text{fp}}, \text{acc}_{\text{fp}} \gets$ \Call{CreateClassifierCAV}{$Z^{-}_{\text{buffer}}, Z^{-}_{\text{train}}$}
        \State \textbf{Compare} $\big(\mathbf{v}_{\text{clf}}, \mathbf{v}_{\text{fp}}\big)$ and $\big(\text{acc}_{\text{clf}}, \text{acc}_{\text{fp}}\big)$
    \end{algorithmic}
\end{algorithm}

\begin{figure}
    \centering
    \begin{subfigure}{0.485\textwidth}
        \centering
        \includegraphics[width=\textwidth]{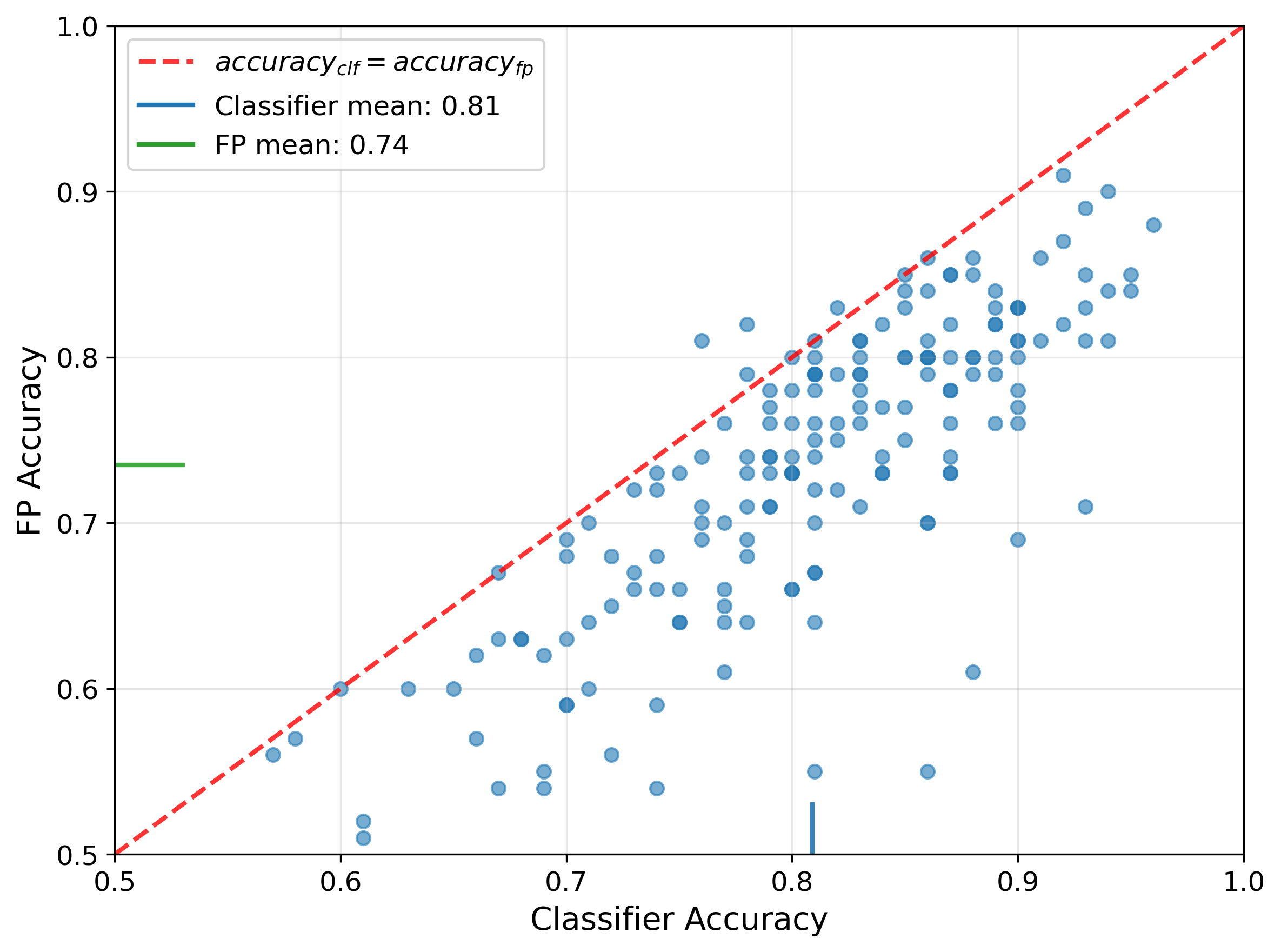}
        \caption{}
        \label{fig:fp_accuracy}
    \end{subfigure}
    \hfill
    \begin{subfigure}{0.48\textwidth}
        \centering
        \includegraphics[width=\textwidth]{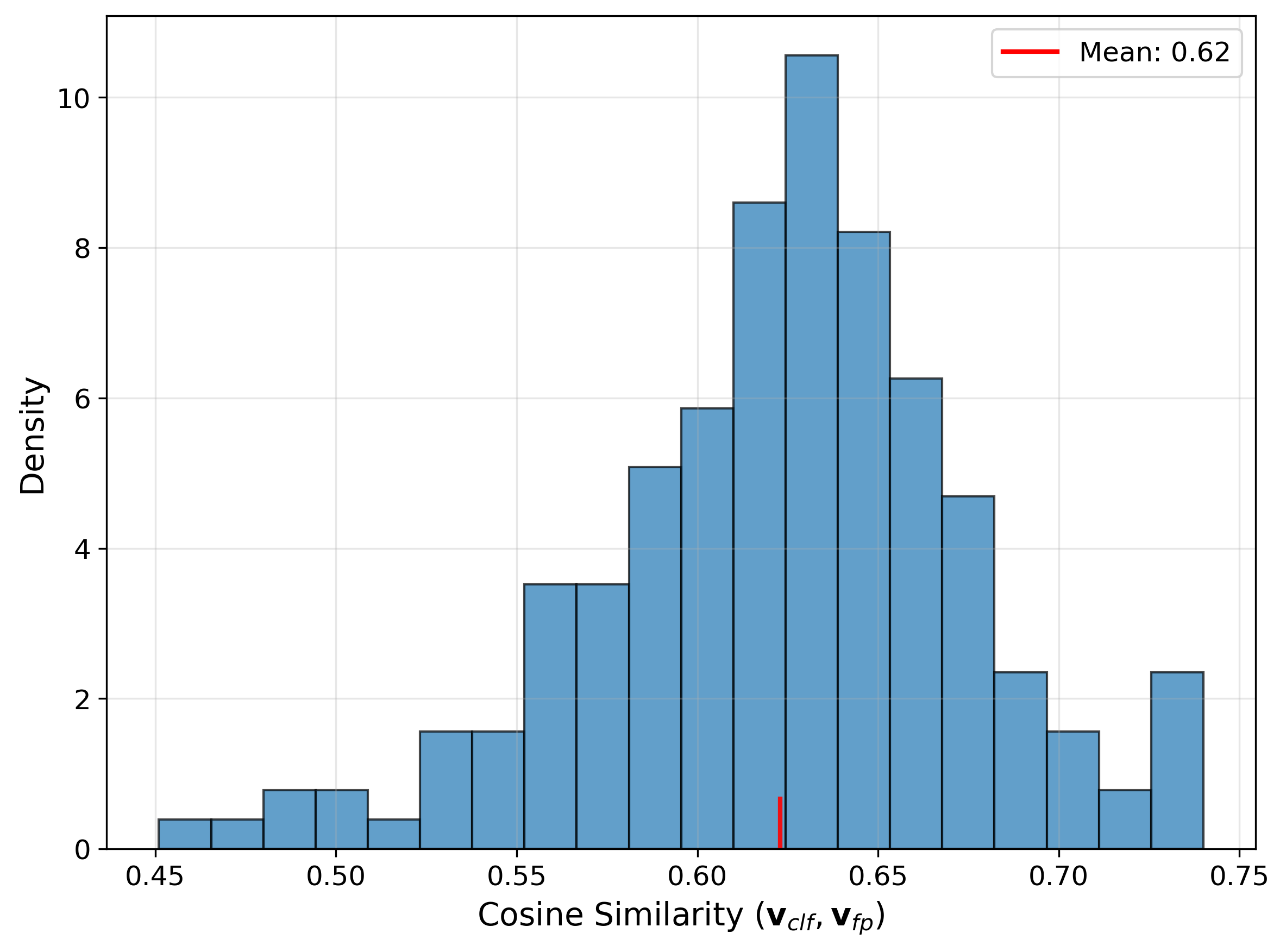}
        \caption{}
        \label{fig:fp_cosim}
    \end{subfigure}
    \caption{Comparison between Classifier- and FP-CAVs trained on the same concepts, with each sample corresponding to a concept. The results show (\subref{fig:fp_accuracy}) similar classification accuracies and (\subref{fig:fp_cosim}) high cosine similarities between Classifier- and FP-CAVs.}
    \label{fig:fp_comparison}
\end{figure}

We find that FP-CAVs, despite being trained without any examples of the target concept, only perform marginally worse than Classifier-CAVs when classifying the same concept. The results reported in Figure~\ref{fig:fp_comparison} show that Classifier-CAVs obtain an average classification accuracy of $81\%$, while FP-CAVs obtain an average accuracy of $74\%$. This relatively high accuracy demonstrates that classification performance alone is not a reliable indicator of concept alignment: high accuracy does not imply that a CAV represents the target concept.

Although FP-CAVs are deliberately misaligned, one might hope that Classifier-CAVs learn to ignore spurious correlations. However, this is likely not the case: The mean cosine similarity between Classifier-CAVs and their corresponding FP-CAVs is $0.62$, indicating substantial overlap in the learned representations. As a baseline for comparison, CAVs trained on different but contextually similar concepts achieve lower similarities, e.g., \texttt{sink} and \texttt{toilet} have a cosine similarity of $0.44$, with more examples in Appendix \ref{ap:clf_cavs_similarity}. The strong similarity between Classifier- and FP-CAVs suggests that a significant portion of their predictive power, and thus classification accuracy, is due to spurious correlations.

\begin{figure}
    \centering
    \includegraphics[width=1\linewidth]{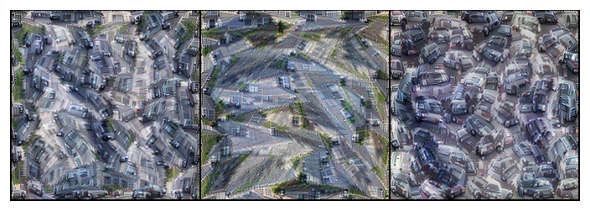}
    \caption{Activation maximization visualizations for Classifier- (left), FP- (middle), and curated CAV (right) for the concept \texttt{car}.}
    \label{fig:cured_car}
\end{figure}

Given the high similarity between Classifier- and FP-CAVs, we further investigate their shared components. To this end, we project $\mathbf{v}_{\text{clf}}$ onto $\mathbf{v}_{\text{fp}}$ and extract the shared component from $\mathbf{v}_{\text{clf}}$. Mathematically, given unit-norm CAVs, this corresponds to rejecting $\mathbf{v}_{\text{fp}}$ from $\mathbf{v}_{\text{clf}}$, computed as
\begin{equation}
    \mathbf{v}_{\text{cured}} = \mathbf{v}_{\text{clf}} - (\mathbf{v}_{\text{fp}} \cdot \mathbf{v}_{\text{clf}}) \mathbf{v}_{\text{fp}} \,.
\end{equation}
Figure~\ref{fig:cured_car} illustrates this process with activation maximization, a visualization technique later described in Section~\ref{sec:activation_maximization}, where $\mathbf{v}_{\text{clf}}$, $\mathbf{v}_{\text{fp}}$, and $\mathbf{v}_{\text{cured}}$ is interpreted. In this example, a Classifier-CAV trained on the concept \texttt{car} appears to activate for vegetation and road segments. After rejecting the FP-CAV, which also shows similar spurious correlations, the cured CAV displays these features far less prominently. This provides further evidence that Classifier-CAVs learn similar spurious correlations to FP-CAVs. Additional results are shown in Appendix~\ref{ap:clf_cavs_false_positives}~and~\ref{ap:clf_cavs_rejection}, showing false positives used to train FP-CAVs and additional examples of rejections. Curating CAVs by rejecting unwanted components, such as those captured by FP-CAVs, presents a promising direction for future research. We suspect that rejecting multiple CAVs, similar to a Gram-Schmidt process, can be useful for aligning CAVs. However, this approach assumes that orthogonalization is more effective than directly training with additional hard negative examples.

\FloatBarrier

\section{Obtaining and visualizing CAVs}\label{sec:obtaining_and_visualizing}
Building on the previous results, we investigate alternative probing methods beyond classifiers and the spurious correlations they tend to capture. We first introduce three additional probing methods in Sections~\ref{sec:pattern_cavs}--\ref{sec:combination_cavs}, followed by a general modification in Section~\ref{sec:translation_invariance} to make them invariant to the position of features. Note that we normalize all CAVs to unit size $\mathbf{v} \leftarrow \frac{\mathbf{v}}{\|\mathbf{v}\|}$ under the assumption that concepts are encoded as directions. To preserve the learned decision boundaries, we also scale the bias terms as $b \leftarrow \frac{b}{\|\mathbf{v}\|}$, where $||\mathbf{v}||$ refers to the pre-normalized magnitude. Finally, in Sections~\ref{sec:prototypical_examples}--\ref{sec:concept_sensitivity}, we apply various feature visualization methods to interpret CAVs, revealing spurious correlations and inductive biases.

\subsection{Pattern-CAVs}\label{sec:pattern_cavs}
\citet{Pahde2025NavigatingDivergence} propose Pattern-CAVs as an alternative to Classifier-CAVs, which are more robust against noise. For binary labels, their method simplifies to the difference between activation centroids
\begin{equation}\label{eq:difference_means}
    \mathbf{v}_{\text{pat}} = \boldsymbol{\mu}^+ - \boldsymbol{\mu}^- \,,
\end{equation}
where the centroids are defined as
\begin{equation}
    \boldsymbol{\mu}^+ = \frac{1}{|Z^+|} \sum_{\mathbf{z}^+ \in Z^+} \mathbf{z}^+ \quad \text{and} \quad \boldsymbol{\mu}^- = \frac{1}{|Z^-|} \sum_{\mathbf{z}^- \in Z^-} \mathbf{z}^- \,.
\end{equation}
To perform binary classification with CAVs without an associated bias term $b$, we find empirically that setting the classification threshold $t$ at the midpoint between the projections onto the centroids
\begin{equation}\label{eq:threshold}
    t = \frac{1}{2} \left( \boldsymbol{\mu}^+ \cdot \mathbf{v} + \boldsymbol{\mu}^- \cdot \mathbf{v} \right) \,,
\end{equation}
result in the highest classification accuracy when using $\mathds{1}[\mathbf{v} \cdot \mathbf{z} > t]$ and $||\mathbf{v}||=1$. We then set $b = -t$ because $\mathds{1}[\mathbf{v}\cdot \mathbf{z} > t]$ and $\mathds{1}[\sigma \left(\mathbf{v} \cdot \mathbf{z} -t \right) > 0.5]$ yield equivalent binary classifications.

\subsection{Segmentation-CAVs}\label{sec:segmentation_cavs}
We hereafter use methods that leverage the spatial structure of activations $\mathbf{z} \in \mathbb{R}^{C \times H \times W}$ in a given layer $l$. Here, $C$ refers to the number of feature maps, or channels, while $H$ and $W$ denote the height and width of the maps, respectively. We follow a similar approach to \citet{Fong2018Net2Vec:Networks}, where probes align feature maps with segmentation masks. We refer to the results of these types of probes as Segmentation-CAVs, learned from a training dataset $\{(\mathbf{x}_i, \mathbf{m}_i)\}_{i=1}^{2N}$ consisting of $N$ positive and $N$ negative examples. Each mask $\mathbf{m}_i \in \{0,1\}^{H \times W}$, obtained by downscaling the segmentation mask to the feature map size, indicates the spatial regions where the concept is present in the corresponding image $\mathbf{x}_i$. To learn Segmentation-CAVs, we first consider how features linearly contribute to the prediction of the classifier $\mathds{1}[\mathbf{v} \cdot \mathbf{z} + b > 0]$. Intuitively, each feature $z_{c,h,w}$ contributes $v_{c,h,w} z_{c,h,w}$ to the prediction, or $\mathbf{v} \odot \mathbf{z}$ in vectorized form, with $\odot$ being the Hadamard product. By summing over channels, we obtain a linear attribution map
\begin{equation}\label{eq:attribution_map}
    \boldsymbol{\phi} = \sum_{c=1}^C \mathbf{v}_{c} \odot \mathbf{z}_{c} \,,
\end{equation}
where $\boldsymbol{\phi} \in \mathbb{R}^{H \times W}$ indicates spatial importance. Subsequently, we learn Segmentation-CAVs $\mathbf{v}_{\text{seg}} \in \mathbb{R}^{C \times H \times W}$ by using attribution maps $\boldsymbol{\phi}$ to predict target masks $\mathbf{m}$. Concretely, we minimize the loss
\begin{equation}\label{eq:segmentation_loss}
    \mathcal{L}_{\text{seg}}(\mathbf{v}) = -\frac{1}{2N H W} \sum_{i=1}^{2N} \left( \mathbf{m}_{i} \cdot \log(\sigma(\boldsymbol{\phi}_{i})) + (\mathbf{1} - \mathbf{m}_{i}) \cdot \log(\mathbf{1} - \sigma(\boldsymbol{\phi}_{i})) \right) \,,
\end{equation}
where the sigmoid function $\sigma$ is used to convert $\boldsymbol{\phi}$ to a map of probabilities. The loss is calculated by taking the mean binary cross-entropy over all $HW$ spatial positions. This encourages $\mathbf{v}_{\text{seg}}$ to represent features that are aligned with the segmentation mask, promoting spatial consistency between features and concepts. Throughout the paper, we use $\cdot$ to denote the standard inner product between CAVs and activations, that is, when both are represented as vectors in $\mathbb{R}^{CHW}$, the notation refers to the dot product, and for tensors it corresponds to the sum of elementwise products. To obtain a bias term associated with $\mathbf{v}_{\text{seg}}$, we repeat the procedure described in Section \ref{sec:pattern_cavs}.

\subsection{Combination-CAVs}\label{sec:combination_cavs}
We find that Segmentation-CAVs often learn different types of features than Classifier-CAVs, see Figures~\ref{fig:car_prototypical}~and~\ref{fig:activation_maximization} for examples. Classifier-CAVs tend to focus on globally predictive features, e.g., wheels for the concept \texttt{car}, while Segmentation-CAVs favor spatially aligned ones, e.g., shiny metal for the same concept. We define Combination-CAVs as directions that integrate both perspectives, aiming to learn features that are both predictive and spatially aligned with the target concept. We refer to the simplest variant as Mixed-CAVs, which is obtained through the convex combination
\begin{equation}
    \mathbf{v}_{\text{mix}} = \beta \mathbf{v}_{\text{clf}} + (1-\beta) \mathbf{v}_{\text{seg}} \,,
\end{equation}
where $\beta \in [0, 1] $ is the weight coefficient, set to $\beta = 0.5$ in our experiments. As an alternative, we explore Joint-CAVs, obtained by jointly optimizing for discriminatory features by Equation~\eqref{eq:classifier_probe} and spatial alignment by Equation~\eqref{eq:segmentation_loss},
\begin{equation}
    \mathcal{L}_{\text{joint}} = \gamma \mathcal{L}_{\text{clf}} + (1-\gamma) \mathcal{L}_{\text{seg}} \,,
\end{equation}
where $\gamma \in [0,1]$ is the weight coefficient. We use $\gamma = 0.99$ in our experiments, but optimal values depend on the specific downstream task and concept. This approach is more flexible than linearly combining CAVs as it can result in CAVs that are not in $\operatorname{span}\{\mathbf{v}_{\text{clf}}, \mathbf{v}_{\text{seg}}\}$. Thus, Joint-CAVs may capture novel features not linearly expressible by the original Classifier- and Segmentation-CAVs.

\subsection{Translation-invariant CAVs}\label{sec:translation_invariance}
We define translation-invariant CAVs as the result of training probes that are invariant to translations in the feature maps $\mathbf{z} \in \mathbb{R}^{C \times H \times W}$. In the literature, classifier probes are often learned by flattening feature maps into vectors in $\mathbb{R}^{CHW}$, allowing the same feature to be weighted differently depending on its spatial position \citep{Nicolson2024ExplainingVectors}. We argue that this approach is inefficient, as it requires more training data to capture the concept across spatial variations. Furthermore, such `position-sensitive' probes are likely more prone to misalignment, as high-dimensional data increase the likelihood of spurious correlations. This motivates the use of probes with constant weights across all spatial positions within each channel. In this section, we derive how Classifier-, Pattern-, Segmentation-, and Combination-CAVs can be learned under the constraint $v_{c,h,w} = \alpha_c$.

The inner product under the translation-invariant constraint can be rewritten as
\begin{equation}
    \mathbf{v} \cdot \mathbf{z} = \sum_{c=1}^C \sum_{h=1}^H \sum_{w=1}^W v_{c, h, w} z_{c, h, w} = \sum_{c=1}^C \alpha_c \sum_{h=1}^H \sum_{w=1}^W z_{c,h,w} \,.
\end{equation}
We define the spatially pooled feature maps as
\begin{equation}
    s_c = \sum_{h=1}^H \sum_{w=1}^W z_{c,h,w} \,,
\end{equation}
and rewrite the inner product to
\begin{equation}\label{eq:gap_innerprod}
    \mathbf{v} \cdot \mathbf{z} = \sum_{c=1}^C \alpha_c s_c = \boldsymbol{\alpha} \cdot \mathbf{s} \,,
\end{equation}
with $\mathbf{s} \in \mathbb{R}^C$ denoting spatially pooled activations, and $\boldsymbol{\alpha} \in \mathbb{R}^C$ denoting learnable weights corresponding to the translation-invariant CAV. Consequently, Classifier- and Segmentation-CAVs can be made translation-invariant by replacing $\mathbf{v} \cdot \mathbf{z}$ in Equations~\eqref{eq:classifier_probe}~and~\eqref{eq:segmentation_loss}, respectively, with $\boldsymbol{\alpha} \cdot \mathbf{s}$. Furthermore, Combination-CAVs are translation-invariant if both constituent loss functions, or the CAVs being combined, are translation-invariant.

For Pattern-CAVs, we similarly pool the feature maps and learn channel weights. The positive and negative means of a spatially pooled channel are defined as
\begin{equation}
    \bar{s}_c^+ = \frac{1}{|Z^+|} \sum_{\mathbf{z}^+ \in Z^+} \sum_{h=1}^H \sum_{w=1}^{W} z^+_{c,h,w} \quad \text{and} \quad \bar{s}_c^- = \frac{1}{|Z^-|} \sum_{\mathbf{z}^- \in Z^-} \sum_{h=1}^H \sum_{w=1}^{W} z^-_{c,h,w} \,.
\end{equation}
The difference between the pooled means is then
\begin{equation}
    \bar{s}_c^+ - \bar{s}_c^- = \sum_{h=1}^H \sum_{w=1}^{W} \left (  \frac{1}{|Z^+|} \sum_{\mathbf{z}^+ \in Z^+}z^+_{c,h,w} - \frac{1}{|Z^-|} \sum_{\mathbf{z}^- \in Z^-} z^-_{c,h,w} \right) =  \sum_{h=1}^H \sum_{w=1}^{W} v_{c,h,w} = HW\alpha_c \,.
\end{equation}

This difference results in a constant weight per channel, since $HW\alpha_c$ is a constant itself. Conveniently, this difference is equivalent to spatially pooling the Pattern-CAV, as defined in Section~\ref{sec:pattern_cavs}. Thus, pooling the activations is equivalent to pooling the Pattern-CAV itself. 

Other pooling strategies that map $\mathbb{R}^{H \times W} \rightarrow \mathbb{R}$, such as Global Average Pooling (GAP) and Global Max Pooling (GMP), also lead to translation-invariant CAVs. For our probing methods, GAP results in identical CAVs, as it only scales the data by some factor that is canceled out during normalization. Using GMP also results in translation-invariant CAVs, but not equivalent to those of GAP. While we do not further experiment with GMP, we expect it to be an efficient method for enforcing both translation and scale invariance. To illustrate, consider a feature map that segments an object: if the object's position and size change, then the GMP value remains unchanged.

Translation-invariant CAVs can be expanded to $\mathbb{R}^{C \times H \times W}$ by the outer product $\mathbf{v} = \boldsymbol{\alpha} \otimes \mathbf{1}_{H \times W}$, ensuring compatibility with methods that expect tensor inputs. For consistency, we continue to represent CAVs in $\mathbb{R}^{C \times H \times W}$ throughout the paper, although using $\boldsymbol{\alpha} \in \mathbb{R}^C$ directly is often approximately $HW$ times more computationally efficient. For instance, classification using $\boldsymbol{\alpha} \cdot \mathbf{s}$ is faster than using $\mathbf{v} \cdot \mathbf{z}$.

\subsection{Prototypical examples}\label{sec:prototypical_examples}
\begin{figure}
    \centering
    \includegraphics[width=1\linewidth]{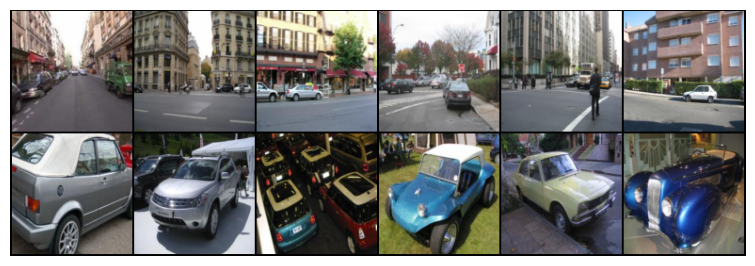}
    \caption{Top-$k$ prototypical examples for the concept \texttt{car} using Classifier- (top row) and Segmentation-CAVs (bottom row). Classifier-CAVs activate for broader scene elements (e.g., roads, vegetation, buildings), while Segmentation-CAVs represent object-specific features (e.g., shiny metal parts).}
    \label{fig:car_prototypical}
\end{figure}

A widely used method for interpreting CAVs is by identifying prototypical examples $\mathbf{x} \in \mathcal{X}$ that yield high similarity between $\mathbf{v}$ and $f_l(\mathbf{x})$. We use test data $|\mathcal{P}_c| = 100$ to sort cosine similarities in descending order and visualize the top-$k$ images in Figure~\ref{fig:car_prototypical}. In the top row, Classifier-CAVs are observed to strongly activate for scenes containing cars, road segments, and buildings for the concept \texttt{car}. In contrast, Segmentation-CAV appears to represent shiny car parts for the same concept. Note the significant difference in car sizes between the methods. This can be explained by the fact that cosine similarity sums over each spatial position. Throughout the feature visualization results, we do not plot Pattern-CAVs as they typically result in indistinguishable visualizations from those of Classifier-CAVs. This is also observed quantitatively, as the average cosine similarity between the these CAVs is surprisingly high (see Appendix~\ref{ap:probing_similarities}).

\subsection{Activation maximization}\label{sec:activation_maximization}
\begin{figure}
    \centering
    \includegraphics[width=1\linewidth]{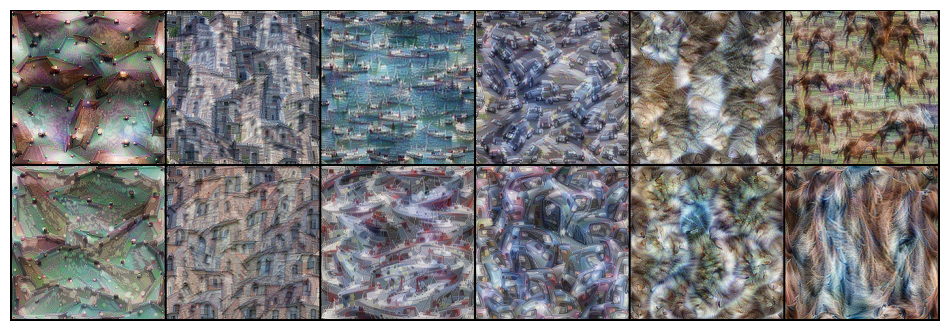}
    \caption{Synthetic images generated via activation maximization for the six concepts \texttt{pool table}, \texttt{building}, \texttt{boat}, \texttt{car}, \texttt{cat}, and \texttt{horse}, as columns, comparing Classifier- (top row) and Segmentation-CAVs (bottom row).}
    \label{fig:activation_maximization}
\end{figure}

To interpret CAVs, it is often useful to optimize randomly initialized images $\mathbf{x}^*$ such that the resulting activations $f_l(\mathbf{x}^*)$ maximize similarity to $\mathbf{v}$. Unlike prototypical examples, activation maximization can generate arbitrary images, unconstrained by the samples in $\mathcal{X}$. Naïvely optimizing for similarity alone, using
\begin{equation}\label{eq:activation_maximization}
    \mathbf{x}^* = \arg\max_{\mathbf{x}} \left( \frac{f_l(\mathbf{x}) \cdot \mathbf{v}}{||f_l(\mathbf{x})||}\right) \,,
\end{equation}
often produces high-frequency artifacts, similar to those seen in adversarial examples. We adopt the method introduced by \citet{olah2017feature}, which combines gradient ascent on Equation~\eqref{eq:activation_maximization} with stochastic transformations applied to $\mathbf{x}$ after each step, regularizing towards natural image statistics. We use the implementation from \citet{greentfrapp_lucent} with the default augmentations, including rotation, scaling, and cropping.

Figure~\ref{fig:activation_maximization} shows synthetic images generated with activation maximization for various CAVs. Although synthetic images can be challenging to interpret, we see exaggerated features highly related to the target concepts, suggesting a degree of alignment. We observe that Classifier-CAVs capture more spurious background features than Segmentation-CAVs, such as boats with less water, cars with fewer roads, and horses with less grass or hay. These visualizations are consistent with the patterns observed in the prototypical examples. Additional results can be seen in Figures~\ref{fig:viz_horse}--\ref{fig:viz_building} in Appendix~\ref{ap:feature_visualization}, showing that activation maximization is largely consistent between CAVs. In general, we observe that Mixed- and Joint-CAVs result in highly similar visualizations, and that translation-invariant CAVs, except for Segmentation-CAVs, result in visualizations similar to their position-sensitive counterparts. Probes trained on the same concept but different training set sizes are also noticeably consistent, especially for Pattern-CAVs, as illustrated in Figure~\ref{fig:activation_maximization_scaling}.

\subsection{Concept localization maps}\label{sec:concept_localization}
\begin{figure}
    \centering
    \begin{subfigure}{\linewidth}
        \centering
        \includegraphics[width=1.0\linewidth]{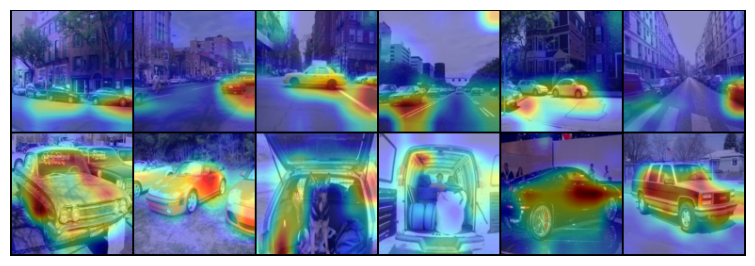}
        \caption{}
        \label{fig:clm_car}
    \end{subfigure}
    
    \begin{subfigure}{\linewidth}
        \centering
        \includegraphics[width=1\linewidth]{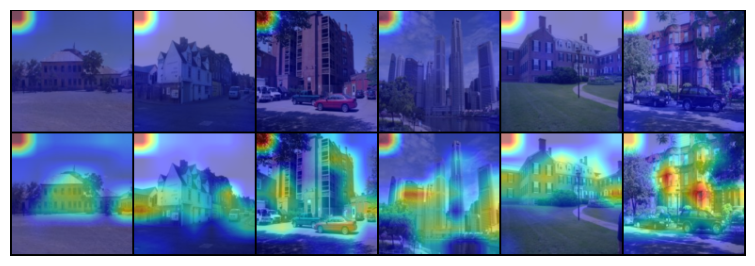}
        \caption{}
        \label{fig:clm_building}
    \end{subfigure}

    \caption{CLMs for the concept (\protect\subref{fig:clm_car}) \texttt{car}, comparing a Classifier-CAV (top row) and Segmentation-CAV (bottom row). For the concept (\protect\subref{fig:clm_building}) \texttt{building}, a Classifier-CAV (top row) can be seen to extensively highlight the upper-left corner of the image, possibly to determine if the sky is present. This tendency is less present for a Classifier-CAV trained with spatial pooling (bottom row).}
    \label{fig:concept_localization_maps}
\end{figure}

Concept localization maps (CLMs) are saliency maps that show which parts of images that CAVs find important for concept classification. Like many saliency methods, CLMs are usually produced by variants of $\texttt{gradient} \times \texttt{input}$, with gradient modifications to account for saturation effects \citep{Shrikumar2017LearningDifferences}. We propose a simpler alternative by shifting the attribution maps defined in Equation~\eqref{eq:attribution_map}. We define shifted attribution maps as
\begin{equation}
    \boldsymbol{\phi}^* = \frac{\mathbf{b}}{HW} + \sum_{c=1}^C \mathbf{v}_{c} \odot \mathbf{z}_{c} \,,
    \label{eq:shifted_attribution_map}
\end{equation}
where the associated bias term is distributed to each spatial position. The reason for including a bias is two-fold. First, since the bias term can dominate concept predictions, the raw attribution map may result in uniform sign values, making it visually uninformative. Second, the distribution ensures that the attribution map sums to the predicted logit, i.e., $\sum_{h,w} \phi^*_{h,w} = \mathbf{v} \cdot \mathbf{z} + b$, thereby satisfying the completeness property. Alternative approaches for incorporating the bias term exist, such as using a mean baseline $\boldsymbol{\phi}^* = \sum_{c=1}^C \mathbf{v}_{c} \odot (\mathbf{z}_{c} - E[\mathbf{z}_{c}])$, as in the original SHAP implementation \citep{Lundberg2017APredictions}. However, this requires a background dataset to calculate expected activations. We further upscale the shifted attribution maps to the original image resolution, apply Gaussian smoothing, and overlay them on the input image. To improve visual clarity, we also apply ReLU to retain only positive attributions.

We plot CLMs for CAVs obtained for the concepts \texttt{car} and \texttt{building} in Figure~\ref{fig:concept_localization_maps}, showing that Classifier- and Segmentation-CAVs find different regions important. Classifier-CAVs often respond to features at the lower regions of cars, such as wheels and road segments, whereas Segmentation-CAVs represent features, often textures, that are more evenly distributed across cars. This interpretation is consistent with observations from the prototypical examples and synthetic images in Figures~\ref{fig:car_prototypical}~and~\ref{fig:activation_maximization}, respectively. Additional results, also for the remaining probes, are provided in Figures~\ref{fig:viz_horse}--\ref{fig:viz_building} in Appendix~\ref{ap:feature_visualization}. For the concepts \texttt{horse} and \texttt{dog}, spatial pooling results in similar CLMs, whereas for the concept \texttt{building}, pooling significantly improves spatial alignment. Visually, Mixed-CAVs seem to be slightly more spatially aligned compared to Joint-CAVs, as seen by less attribution to background features.

Interestingly, we observe that for some concepts, such as \texttt{building}, CAVs frequently learn a feature in the upper corners of the images, refer to Appendix~\ref{ap:clf_cavs_false_positives}~and~\ref{ap:feature_visualization} for further examples. We suspect that information in the upper corners serves as a heuristic for detecting whether an image was taken outdoors, possibly by checking for the presence of sky. This spurious correlation is subtle and not easily detected by other visualization methods, highlighting the unique diagnostic value of our CLM method. We further find that translation-invariant probes are significantly less prone to learning this specific pattern. Yet its persistence is surprising: if a sky detector were truly present, one would expect the entire sky region to be highlighted, not just one specific position. We suspect this behavior arises from boundary effects introduced by convolutional padding, which enables CNNs to encode absolute positional information \citep{Kayhan2020OnLocation}.

\subsection{Concept sensitivity}\label{sec:concept_sensitivity}
\begin{figure}
    \centering
    \includegraphics[width=1\linewidth]{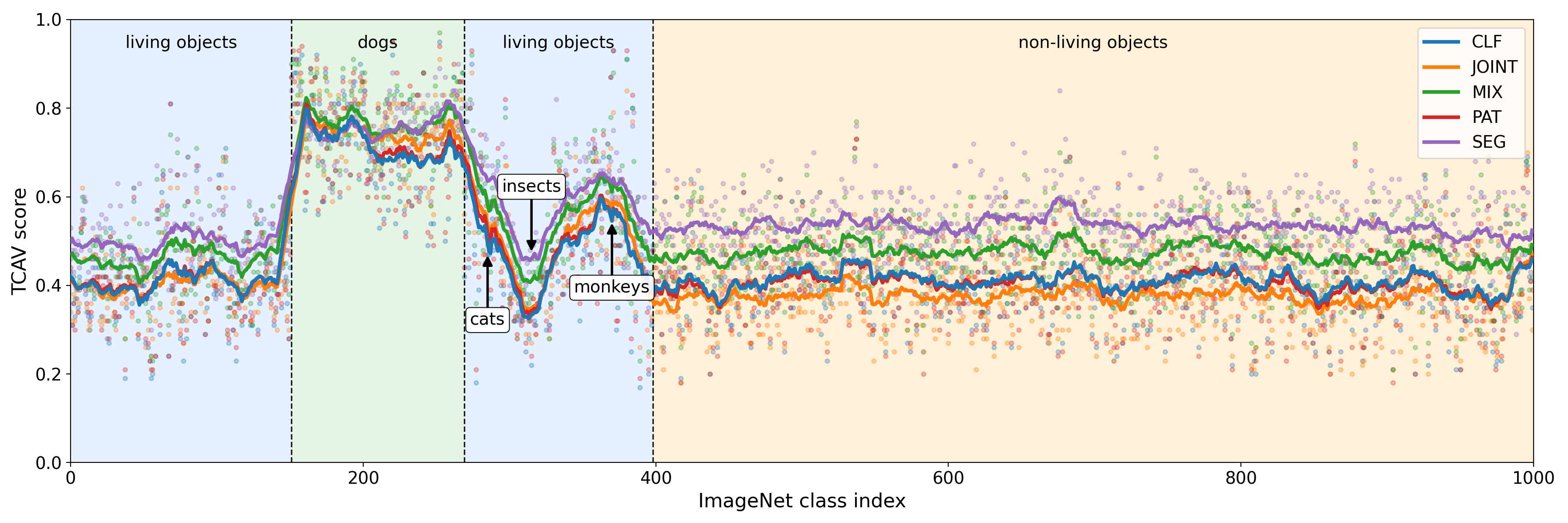}
    \caption{$\text{TCAV}$ scores for the concept \texttt{dog} across all $1000$ ImageNet classes, using different probing methods. Semantic ordering of class indices reveals clear positive sensitivity to the dog classes (indices $152$-$269$).}
    \label{fig:concept_sensitivity_dog}
\end{figure}

To further assess concept alignment, we test whether perturbing activations in the direction of a CAV produces the expected changes in model predictions. For state-of-the-art models like ResNet50, trained to classify $1000$ classes, of which $117$ are different dog breeds, we assume the existence of an abstract \texttt{dog} direction that should positively influence dog classes more than others. Following \citet{Kim2018InterpretabilityTCAV}, we compute directional derivatives as
\begin{equation}
    S_{c,l,k}(\mathbf{x}) = \nabla f_{l \rightarrow L, k}(f_l(\mathbf{x})) \cdot \mathbf{v} \,,
\end{equation}
where $f_l(\mathbf{\cdot})$ maps inputs to intermediate activations at layer $l$, and $f_{l \rightarrow L, k}(\cdot)$ maps those activations to the final layer $L$ with predicted probability of class $k$. We then calculate the fraction of positive concept sensitivity by
\begin{equation}
    \text{TCAV}_{c,l,k} = \frac{|\mathbf{x} \in \mathcal{P}_c : S_{c,l,k}(\mathbf{x}) > 0|}{|\mathcal{P}_c|} \,.
\end{equation}
We calculate $\text{TCAV}_{c,l,k}$ for the $1000$ classes that ResNet50 is trained on, with \texttt{dog} as concept $c$. Since the ImageNet classes are semantically ordered, we can compactly visualize all TCAV scores as in Figure~\ref{fig:concept_sensitivity_dog}, where each point represents a single $\text{TCAV}$ score and the lines indicate trends via moving averages. For example, the first $400$ classes are living objects, which are further divided into specific categories such as different dog breeds (indices $152$-$269$) and cat breeds (indices $282$-$294$). Following the moving average lines, we see that the classes corresponding to dog breeds are positively sensitive to the \texttt{dog} direction as found by all probing methods. That is, moving towards the dog direction to activations increases the prediction of dog classes about $70\%$ of the time. If the concept is irrelevant for a class, we expect TCAV scores close to $50\%$, which is mostly the case for the other classes. We also observe a strong correlation between TCAV scores for the different probing methods, as seen in local variations in sensitivity. In fact, most of the differences in TCAV scores seem to be explained by a small constant shift. This suggests that, under our assumptions, all probing methods non-trivially capture the target concept, but with intrinsic differences due to their different inductive biases. 

Repeating the $\text{TCAV}$ experiment with translation-invariant CAVs, presented in Appendix \ref{ap:tcav}, results in similar but interesting observations. We no longer observe a strong constant shift between methods, as all translation-invariant CAVs result in remarkably similar TCAV scores. This is likely due to an effect caused by spatial pooling, which produces a subspace in the solution space where probes are more likely to result in similar CAVs. Furthermore, the TCAV scores are more centered at $50\%$, and the spikes for correlated concepts, e.g., primates around indices $366$-$383$ are less significant, especially for the Classifier- and Segmentation-CAVs. These results indicate that GAP has a positive effect for TCAV scores as an explanation method.

Additional experiments also indicate that TCAV scores are highly dependent on the selected layer to probe. While most layers yield similar results, we find multiple late layers that do not show a significant spike for dog classes. Importantly, this likely reflects poor concept alignment rather than poor linear separability. As noted in prior work \citep{Alain2017UNDERSTANDINGPROBES, Kim2018InterpretabilityTCAV, McGrath2022AcquisitionAlphaZero}, linear availability tends to strictly increase throughout model layers. We also tested for increasing the amount of training data, but it only provided a slight improvement in $\text{TCAV}$ scores across all methods except for Pattern-CAVs. Thus, we highlight the importance of directly evaluating concept alignment when selecting layers to probe, rather than relying on classification accuracy alone.

\FloatBarrier
\section{Quantifying concept alignment}\label{sec:quantifying_alignment}
In this section, we move away from visual interpretations of CAVs and introduce three quantitative metrics for assessing concept alignment. We provide a comparison of these metrics across the probing methods described in Sections~ \ref{sec:linear_clf_probes}~and~\ref{sec:pattern_cavs}-\ref{sec:translation_invariance}. We also analyze how concept alignment scales with the size of the concept training dataset.

\subsection{Normal and hard accuracy}
As demonstrated in Section~\ref{sec:fp_cavs}, probe classification accuracy can be misleading due to spurious correlations, making it an unreliable standalone metric for concept alignment. We take inspiration from distributionally robust optimization and test for a simplified version of worst-group accuracy \citep{Sagawa2020DistributionallyGeneralization}. To illustrate the importance of worst-group accuracy, consider that most images with the concept \texttt{horse} are taken outdoors on pastures. Probes may rely on pasture as a spurious cue, succeeding on typical horse images but failing on atypical ones. If the probe strongly relies on spurious correlations, it will fail to classify groups where the spurious correlation is absent. This motivates evaluating the accuracy on hard groups, such as images of horses on beaches, since high performance on these examples implies that the probe has learned robust correlations. To this end, we replace the backgrounds in the test data with random images that do not contain the concept, and we interpret the dataset as containing rare but plausible examples of the concept where background features are no longer useful. We refer to the classification accuracy on test images with randomized backgrounds as \emph{hard accuracy} to distinguish it from normal accuracy evaluated on unaltered images. If probes rely on spurious background correlations, we expect hard accuracy to be significantly lower than the normal accuracy. Note that hard groups need not be defined in terms of the background. However, we investigate spurious background correlations due to the prevalence of this vulnerability as revealed by feature visualization methods.

In Table~\ref{tab:alignment_metrics}, we see that hard accuracies are significantly lower than the corresponding normal accuracies for all probing methods. Although Segmentation-CAVs exhibit relatively low normal accuracy, their performance remains stable on randomized backgrounds, yielding the highest overall hard accuracy. Combination-CAVs seem to strike a balance between the two, achieving relatively high normal and hard accuracy. Translation-invariant CAVs, having $HW$ fewer parameters, still obtain similar accuracies as their counterparts, if not better. Furthermore, Figure~\ref{fig:alignment_metrics_scaling} shows that both Classifier- and Segmentation-CAVs consistently improve in classification performance as the training data size increases. In contrast, Pattern-CAVs are accurate with small dataset sizes, but they do not improve when provided additional training data. This is likely due to the nature of averaging, which limits the influence of individual counterexamples. For instance, if a spurious feature $z_{\text{sp}}$ appears in $99$ out of $100$ positive samples, its influence on the Pattern-CAV remains strong, even if the single counterexample contradicts it. In contrast, classifier-based probes can flexibly adjust their decision boundaries when exposed to diverse data, allowing them to reduce reliance on spurious correlations as the correlations become less predictive.

\subsection{Segmentation score}\label{sec:segmentation_score}
In this section, we quantify the extent to which the concept logit is attributed to regions where the concept is actually present, and consequently, how much is due to the background. Using shifted attribution maps $\boldsymbol{\phi}^*$, as defined in Equation~\eqref{eq:shifted_attribution_map}, the logit $\mathbf{v} \cdot \mathbf{z} + b$ is decomposed into individual spatial contributions. Using this decomposition, we define the segmentation metric as
\begin{equation}
    S_c = \frac{1}{N} \sum_{i=1}^N \frac{\mathbf{m}_i \cdot \boldsymbol{\phi}_i^+}{\sum_{h,w}\phi_{i,h,w}^+} \,,
\end{equation}
with $\boldsymbol{\phi}_i^+ = \text{ReLU}(\boldsymbol{\phi}^*_i)$ being the spatial attribution map with positive contributions to the logit, and $\mathbf{m}_i$ denoting the segmentation mask. The numerator increases when $m_{i,h,w} = 1$ and $\phi^+_{i,h,w} > 0$, valuing correctly assigned positive attributions, while the denominator normalizes the score. Intuitively, the score represents the fraction of positive attributions that fall within the target segmentation mask. An important property of this metric is that normalization ensures that the value of the classification logit does not matter, only how it is spatially distributed. This property is important for making fair comparisons, as the distribution of logits largely depend on the probing method.

As expected, Segmentation-CAVs achieve the highest segmentation scores, as seen in Table~\ref{tab:alignment_metrics}, followed by Combination-CAVs, which are also obtained by segmentation as well as classification. Translation-invariant CAVs consistently outperform their position-sensitive counterparts. This likely indicates that concepts are encoded at the channel level, rather than as combinations of channel and position. Since CNNs are approximately translation-equivariant \citep{Biscione2021ConvolutionalBe}, i.e., shifting the input leads to similarly shifted activations, linear combinations of channels should best represent concepts. However, because CNNs are not perfectly translation-equivariant, it remains plausible that they also encode concepts using absolute positions. Similarly, some features are sparse at specific positions, which may encourage the model to encode features in superposition. To ensure accurate probing of concepts, future work should investigate the extent to which CNNs encode the absolute position of features.

\subsection{Augmentation robustness}
Let the concept function $p_c(\mathbf{x}) \in \{{ 0,1 \}}$ denote whether the concept $c$ is present in an image $\mathbf{x}$, returning $1$ if present and $0$ otherwise. For many concepts, there exist transformations $T_c : \mathcal{X} \rightarrow \mathcal{X}^*$ that produce realistic images within the same domain such that $p_c(\mathbf{x}) = p_c(T_c(\mathbf{x}))$. For example, by flipping an image horizontally, concepts under the category of objects do not cease to exist or suddenly appear, but they remain unchanged. Since $p_c(\mathbf{x}) = p_c(T_c(\mathbf{x}))$ holds for the true concept function, this invariance should ideally be preserved by probes $\mathds{1}\left[ \mathbf{v} \cdot f_l(\mathbf{x}) + b\right] \approx p_c(T_c(\mathbf{x}))$. While binary predictions could be compared before and after transformation, they obscure meaningful changes in activation. For example, predictions near $1.0$ and $0.5$ may both result in the same binary label, even though they differ substantially in confidence. Instead, we measure the absolute difference in dot products $|\mathbf{v} \cdot f_l(\mathbf{x}) - \mathbf{v} \cdot f_l(T_c(\mathbf{x}))|$. To create a bounded metric, we define the change in activations as
\begin{equation}
    \Delta \mathbf{z} = f_l(\mathbf{x})- f_l(T_c(\mathbf{x})) \,,
\end{equation}
and factorize the difference in dot products to $|\mathbf{v} \cdot \Delta \mathbf{z}|$. By the Cauchy-Schwarz inequality and unit size CAVs, we obtain the upper bound $|\mathbf{v} \cdot \Delta \mathbf{z}| \leq ||\Delta \mathbf{z}||$. Thus, we define the augmentation robustness metric as
\begin{equation}
    R_c = 1- \frac{1}{N} \sum_{i=1}^N \frac{|\mathbf{v} \cdot \Delta \mathbf{z}_i|}{||\Delta \mathbf{z}_i||} \,,
\end{equation}
where $R_c = 1$ represents perfect augmentation robustness. This occurs when $|\mathbf{v} \cdot \Delta \mathbf{z}| = 0$, implying that the change vector is orthogonal to the CAV. Conversely, the worst possible score $R_c = 0$ occurs when $\mathbf{v}$ and $\Delta \mathbf{z}$ are collinear, i.e., when the transformation alters activations only along the CAV direction.

\begin{figure}
        \centering
        \includegraphics[width=0.95\linewidth]{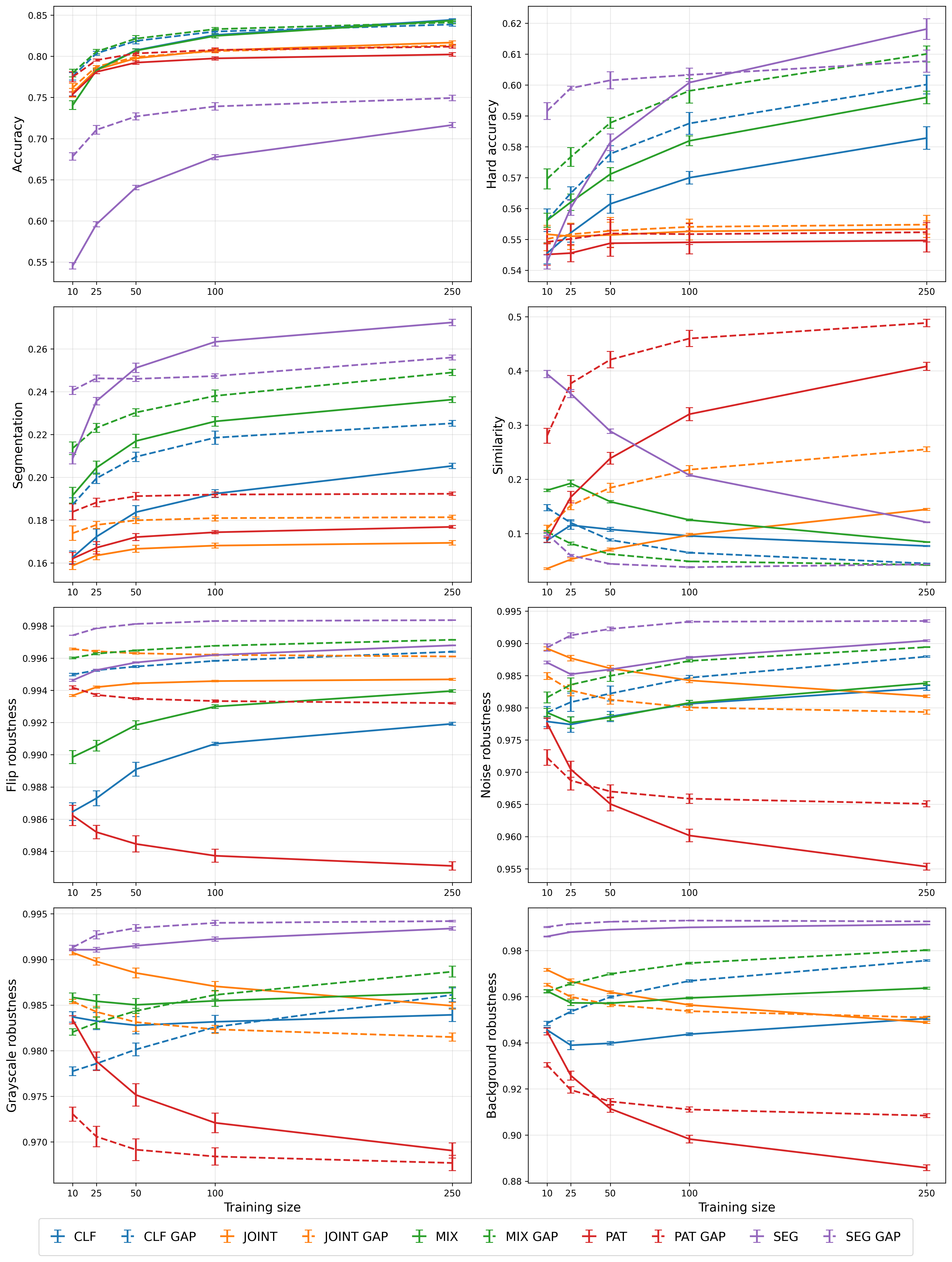}
        \caption{Concept alignment metrics for different probing methods trained across $N=\{10, 25, 50, 100, 250\}$ sizes of training data. All concepts with at least 350 examples are tested and averaged. The experiment is repeated $5$ times, and the standard deviation is shown as error bars.}
        \label{fig:alignment_metrics_scaling}
\end{figure}

We evaluate robustness using four transformations: horizontal flipping, grayscaling, Gaussian noise, and random background replacement. The results presented in Figure~\ref{fig:alignment_metrics_scaling} show high robustness values for all transformations, and that scores tend to improve as the training data increase. Spatial pooling is particularly effective for obtaining invariance to these transformations, especially horizontal flipping. Interestingly, Pattern- and Joint-CAVs exhibit a decline in robustness as the training data increases, despite maintaining stable classification and segmentation performance. To investigate this, we introduce a \emph{Similarity} metric defined as the average cosine similarity between CAVs for all probed concepts. Ideally, CAVs should be orthogonal to ensure disentangled representations. While low Similarity scores do not guarantee that a probing method produces aligned CAVs, as most directions in high-dimensional space are approximately orthogonal, high Similarity scores likely imply misalignment. We observe that Similarity tends to be inversely related to robustness, and that Pattern- and Joint-CAVs increase in Similarity as training data grows. Methods with high Similarity scores are more likely to learn CAVs that lie within high-density regions of the activation space, i.e., along manifolds. As a result, augmentations have more concepts to interfere with. In contrast, the other methods tend to learn more disentangled representations as training data increases, which in turn leads to fewer components that the augmentations can interfere with.

\input{metrics_table}

\FloatBarrier
\section{Discussion}\label{sec:discussion}
Our findings have several important implications for concept-based explanation methods: We demonstrate that both existing and novel probing methods result in misaligned CAVs. Using FP-CAVs, we show that probes can achieve high classification accuracy while relying entirely on spurious correlations. These two findings are of significant concern as misaligned CAVs can lead to misleading explanations in downstream applications. For instance, if a CAV represents both the target concept and confounding features, it becomes unclear whether the concept itself or its confounders dominate when using explanation methods like TCAV scores. This ambiguity undermines the trustworthiness of concept-based explanations. At present, due to the unreliably of classification accuracy, no standard method exists for assessing concept misalignment. However, our results demonstrate the usefulness of various feature visualization methods and alignment metrics, which we hope will be adopted and inform future methodological development.

The probing methods outlined in Sections~\ref{sec:pattern_cavs}--\ref{sec:combination_cavs} show different strengths and limitations, making them suitable for different use cases. For example, Pattern-CAVs are preferable in low training data settings, Classifier-CAVs when concept training data is abundant, Segmentation-CAVs when probing object-like concepts with available segmentation masks, and Combination-CAVs when it is crucial to balance discriminatory and spatially aligned features. Our analysis suggests that most probing methods produce CAVs with nuanced differences, and the choice of layer might be equally important. While probes in later layers tend to achieve higher classification accuracy, they may result in worse concept alignment. This challenges the common practice of selecting late layers based on accuracy, highlighting the need for alignment-based layer selection. However, further analysis on the relationship between alignment and layers is necessary. Notably, none of the trained probes achieved full alignment -- which may be impossible to achieve in practice, given that some concepts are not linearly represented within the target layer -- further highlighting the importance of complementary evaluation metrics for characterizing probe behavior and alignment.

Our analysis and the aforementioned insights were achieved using a toolbox of feature visualization techniques. In particular, activation maximization and our proposed CLM method consistently revealed spurious correlations across all experiments and probing methods. CLMs proved particularly useful for uncovering subtle spurious correlations, as exemplified by the frequent reliance on a feature in the upper corners, likely used to detect the presence of sky. Our CLM approach, based on spatial attribution and bias distribution, is straightforward to implement and provides valuable diagnostics for concept alignment. In contrast, prototypical examples provided an overview of what CAVs represent, but proved less effective for identifying spurious correlations. These visual interpretations are further backed up by quantitative results. As demonstrated, feature visualization 
is a useful tool for assessing alignment, and we recommend that researchers utilize it to verify their probes.

Spatial pooling, segmentation, and increasing training data size improved concept alignment across most metrics and probes. Larger concept training datasets lead to increasingly discriminative features, as evidenced by improved accuracy and changes in activation maximization. However, Pattern- and Joint-CAVs were less impacted by the size of the concept training dataset. For Pattern-CAVs, this is likely due to their reliance on averaging, which leads to fast convergence and insensitivity to hard data instances. Consequently, this makes them a viable option when concept training data is sparse. Moreover, Pattern-CAVs may be particularly useful for tasks such as model debiasing, given that they capture all features correlated with the target concept. Joint-CAVs, while conceptually appealing, proved difficult to optimize reliably across all concepts. Despite this, we believe they hold promise if optimization stability can be improved. In the meantime, the method producing Mixed-CAVs proved more consistent and reliable for interpolating between spatially aligned and discriminative features. Segmentation-CAVs achieved strong alignment but require segmentation masks and tend to favor particular features, such as textures, rather than discriminatory features that reflect the essence of a concept, limiting their applicability to specific object concepts.

By incorporating prior knowledge about the network architecture and target concept into the probe, we regularize the solution space towards preferred CAVs that optimizers may otherwise not find. This is particularly effective for Classifier-CAVs, as the solution space of hyperplanes that obtain perfect linear separation is enormous due to the high dimensionality of the activation space. Spatial pooling can be interpreted as narrowing the solution space to a subspace where all candidate CAVs satisfy the restriction and still obtain perfect separation. Similarly, GMP can be used to introduce size invariance in addition to translation invariance. While spatial pooling is specialized for CNNs, training probes with augmented inputs is a general strategy that can be applied to most domains and architectures. For example, language model probes could be trained to be invariant to spelling errors, capitalization, or even the language used. While we have applied augmentations to assess robustness, it could alternatively be used during training to encourage invariance. However, if a method is trained and evaluated on the same augmentation, the corresponding metric loses its diagnostic potential.

A few comments regarding the scope of our work are in order: Studying \emph{linear} probes implies the assumption that concepts are encoded as directions in activation space. While many concepts may be better captured through non-linear probes, such probes are likely more difficult to interpret and evaluate. Furthermore, our work probes segmentable objects, implicitly assuming that concepts have absolute positions in an image. While Classifier- and Pattern-CAVs can be applied to non-spatial concepts, e.g., \texttt{grayscaled image}, such concepts are not suited for Segmentation-CAVs.

Our findings suggest several promising directions for future work. First, data quality manifests as a critical factor in concept alignment. Techniques such as hard negative mining or targeted data augmentation may improve probe robustness. Second, concept curation via orthogonalization, such as rejecting FP-CAVs using a Gram-Schmidt process, could help remove spurious components. However, fully optimizing for orthogonality, e.g., by penalizing cosine similarity during training, may be impractical in high-dimensional spaces, where the number of approximately orthogonal vectors grows exponentially. In this scenario, orthogonality regularization likely requires training many CAVs jointly to sufficiently constrain the solution space. Thus, finding disentangled representations in the form of CAVs might be more reliable when using unsupervised methods. Third, our finding that spurious correlations in the upper corners persist for translation-invariant CAVs suggests that ResNet50 encodes positional information through boundary effects introduced by convolutional padding. Investigating whether CNNs represent features at absolute positions could yield deeper insights into how concepts are represented and how probes can be made more reliable. Finally, extending our analysis to include layer selection could reveal how concept alignment varies across the network and the extent it correlates with classification accuracy. If discrepancies are found, then new strategies should be developed for selecting target layers for probing.

\FloatBarrier
\section{Conclusion}\label{sec:conclusion}
Concept alignment is crucial for assuring effective and trustworthy use of concept-based explanations. We demonstrate that the probe's classification accuracy is unreliable for assessing concept alignment and that probes often achieve high accuracy by heavily relying on spurious correlations. Through a combination of visualization techniques and quantitative metrics, we show how different probing methods vary in their ability to capture the target concept. Feature visualization methods, including our proposed CLM method, along with metrics such as hard accuracy, segmentation score, and augmentation robustness, provide practical tools for evaluating alignment beyond classification accuracy. We find that translation-invariant and segmentation-based methods consistently improve alignment, and our results underscore the importance of creating specialized probes by leveraging knowledge about the model architecture and target concepts. Our findings highlight the necessity of evaluating concept alignment and highlight the limitations of current probing methods, thereby paving the way for more reliable and robust concept-based explanations.

\clearpage
\bibliographystyle{apalike}
\bibliography{references}

\clearpage

\appendix

\section{Classifier-CAVs}\label{ap:classifier_cavs}

\subsection{CAV distributions}\label{ap:clf_cavs_distributions}
\begin{figure}[H]
    \centering
    \includegraphics[width=1\linewidth]{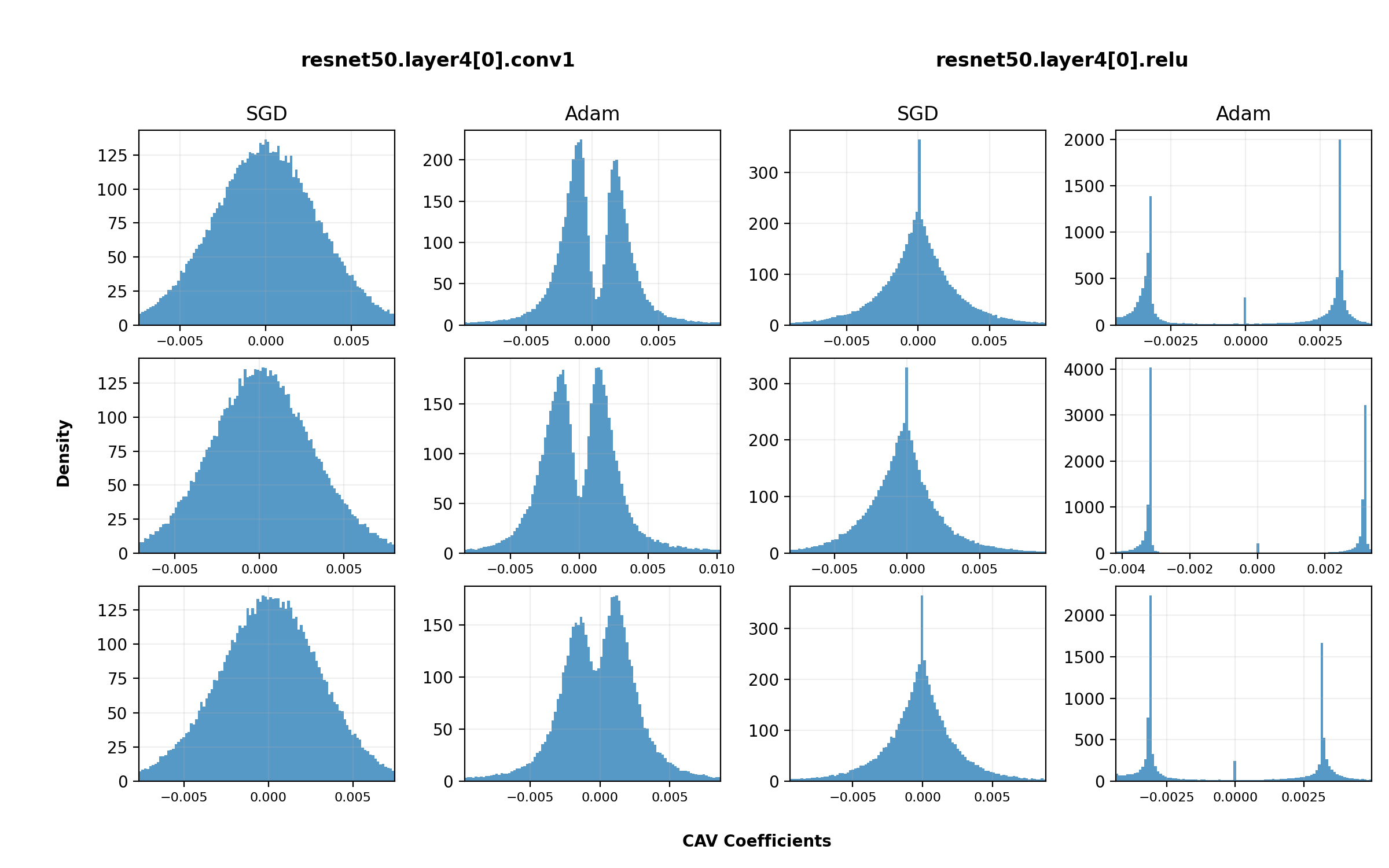}
    \caption{Distributions of weights for Classifier-CAVs trained on the concepts \texttt{flower}, \texttt{horse}, and \texttt{grass} (as rows, respectively). The CAVs are obtained from pre-ReLU (columns 1 and 2) and post-ReLU (columns 2 and 3) activations. Classifier-CAVs obtained with SGD tend to learn approximately normally distributed weights for pre-ReLU activations and Laplace distributed with excessive zeros for post-ReLU activations. In contrast, Adam tends to learn bimodal distributions. This tendency is consistent across most concepts and later layers.}
    \label{fig:clf_cav_distributions}
\end{figure}

\subsection{Similarity between CAVs}\label{ap:clf_cavs_similarity}
\begin{table}[H]
    \centering
    \parbox{0.4\linewidth}{
        \caption{Sorted cosine similarities between Classifier-CAVs trained on different concepts}
        \centering
        \begin{tabular}{ccc}
        \hline
    \addlinespace
    Concept 1 & Concept 2 & Cosine similarity $\downarrow$\\
    \hline
    \addlinespace
    
    countertop & toilet & 0.535 \\
    towel & toilet & 0.502 \\
    countertop & screen door & 0.484 \\
    bush & river & 0.448 \\
    sink & toilet & 0.443 \\
    
    $\vdots$ & $\vdots$ & $\vdots$ \\

    sky & wardrobe & -0.343 \\
    sky & towel & -0.346 \\
    sky & basket & -0.355 \\
    sky & countertop & -0.360 \\
    sky & toilet & -0.374 \\
    
    \hline
    \end{tabular}
    \label{tab:clf_cav_similarity}
    }
\end{table}

\subsection{False positives}\label{ap:clf_cavs_false_positives}
\begin{figure}[H]
    \centering
    \begin{subfigure}{\linewidth}
        \centering
        \includegraphics[width=1.0\linewidth]{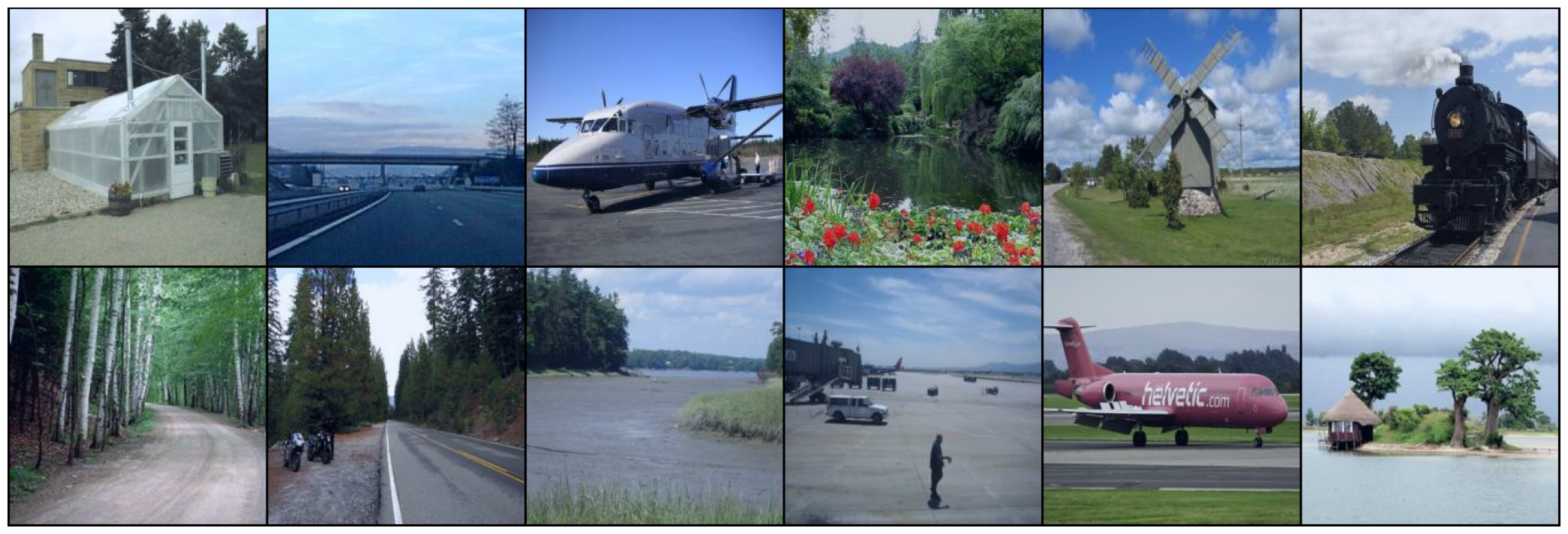}
        \caption{}
        \label{fig:fp_building}
    \end{subfigure}

    \begin{subfigure}{\linewidth}
        \centering
        \includegraphics[width=1.0\linewidth]{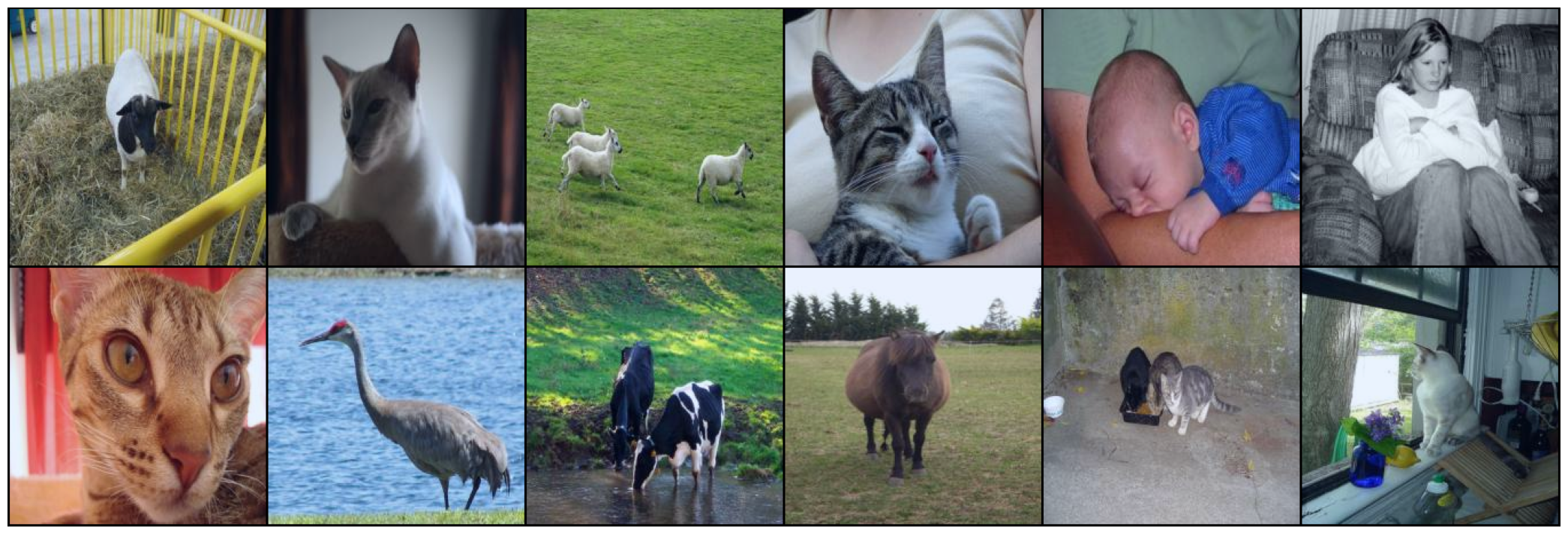}
        \caption{}
        \label{fig:fp_dog}
    \end{subfigure}
    
    \begin{subfigure}{\linewidth}
        \centering
        \includegraphics[width=1.0\linewidth]{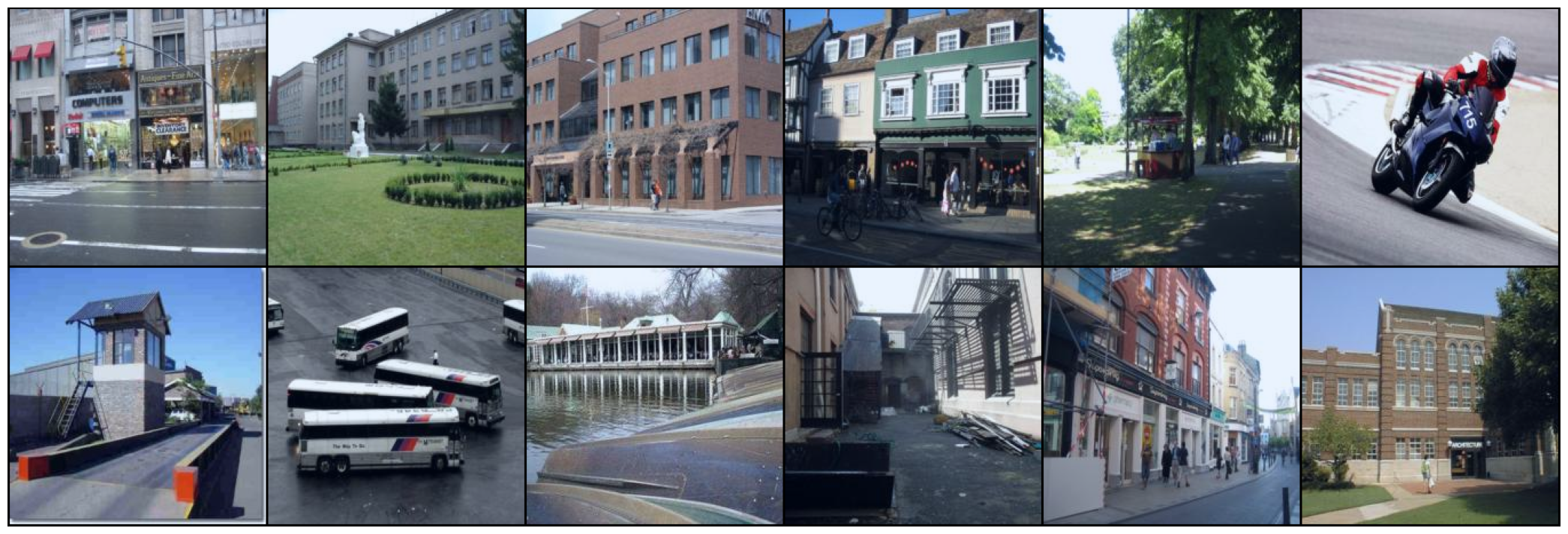}
        \caption{}
        \label{fig:fp_car}
    \end{subfigure}

    \caption{False positive images collected from Classifier-CAVs for concepts (\protect\subref{fig:fp_building}) \texttt{building}, (\protect\subref{fig:fp_dog}) \texttt{dog}, and (\protect\subref{fig:fp_car}) \texttt{car}. The sky seems important for classifying \texttt{buildings}, living animal features are likely important for \texttt{dogs}, and roads and buildings are associated with \texttt{car}.}
    \label{fig:false_positives}
\end{figure}

\subsection{Rejecting FP-CAVs}\label{ap:clf_cavs_rejection}
\begin{figure}[H]
    \centering
    \begin{subfigure}{\linewidth}
        \centering
        \includegraphics[width=1.0\linewidth]{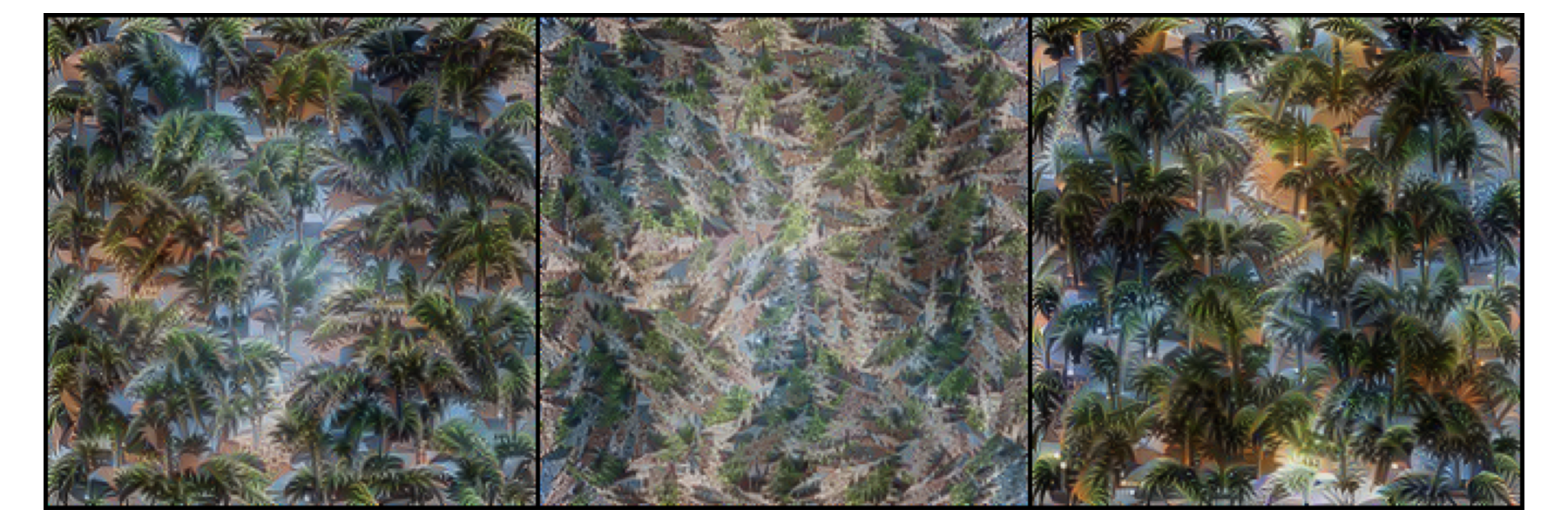}
        \caption{}
        \label{fig:cured_palm}
    \end{subfigure}

    \begin{subfigure}{\linewidth}
        \centering
        \includegraphics[width=1.0\linewidth]{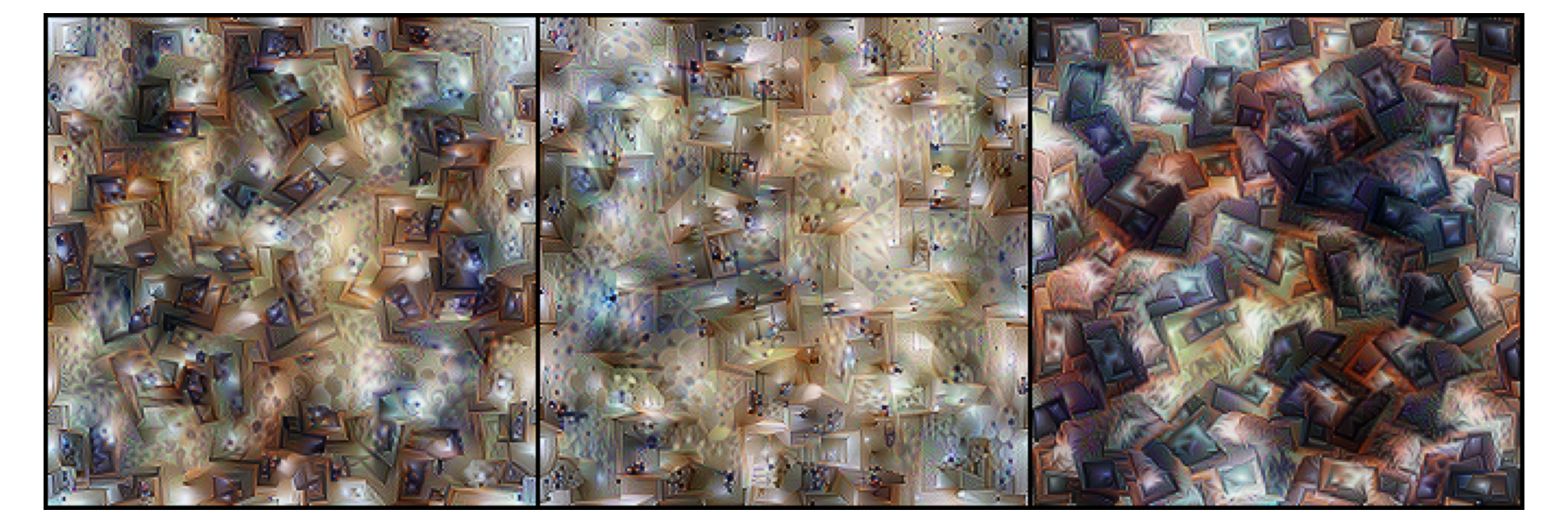}
        \caption{}
        \label{fig:cured_television}
    \end{subfigure}
    
    \begin{subfigure}{\linewidth}
        \centering
        \includegraphics[width=1.0\linewidth]{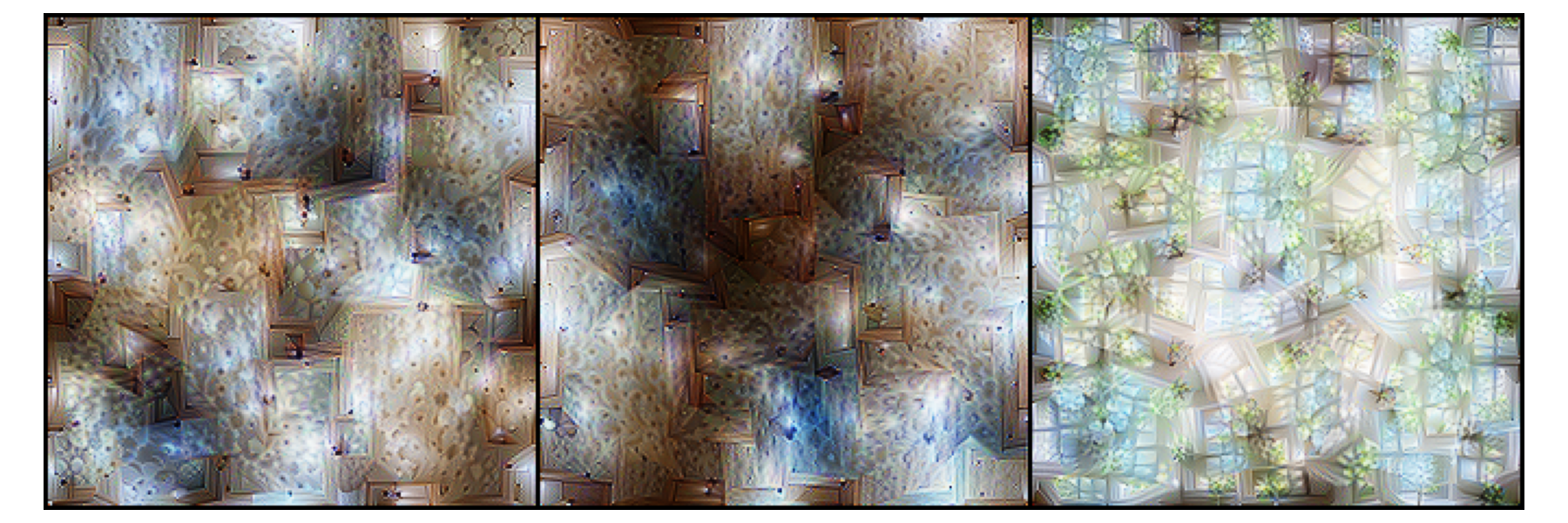}
        \caption{}
        \label{fig:cured_windowpane}
    \end{subfigure}

    \caption{Activation maximization visualizations of CAVs where Classifier-CAVs are plotted in the left column, corresponding FP-CAVs in the middle, and the rejections to the right. The CAVs are trained on the concepts (\protect\subref{fig:cured_palm}) \texttt{palm tree}, (\protect\subref{fig:cured_television}) \texttt{television}, and (\protect\subref{fig:cured_windowpane}) \texttt{windowpane}. For most CAVs, rejecting an associated FP-CAV tends to result in clearer visualizations of the target concept.}
    \label{fig:appendix_cured}
\end{figure}

\section{Probing method similarities}\label{ap:probing_similarities}
\begin{figure}[H]
    \begin{subfigure}{\linewidth}
        \centering
        \includegraphics[width=0.7\linewidth]{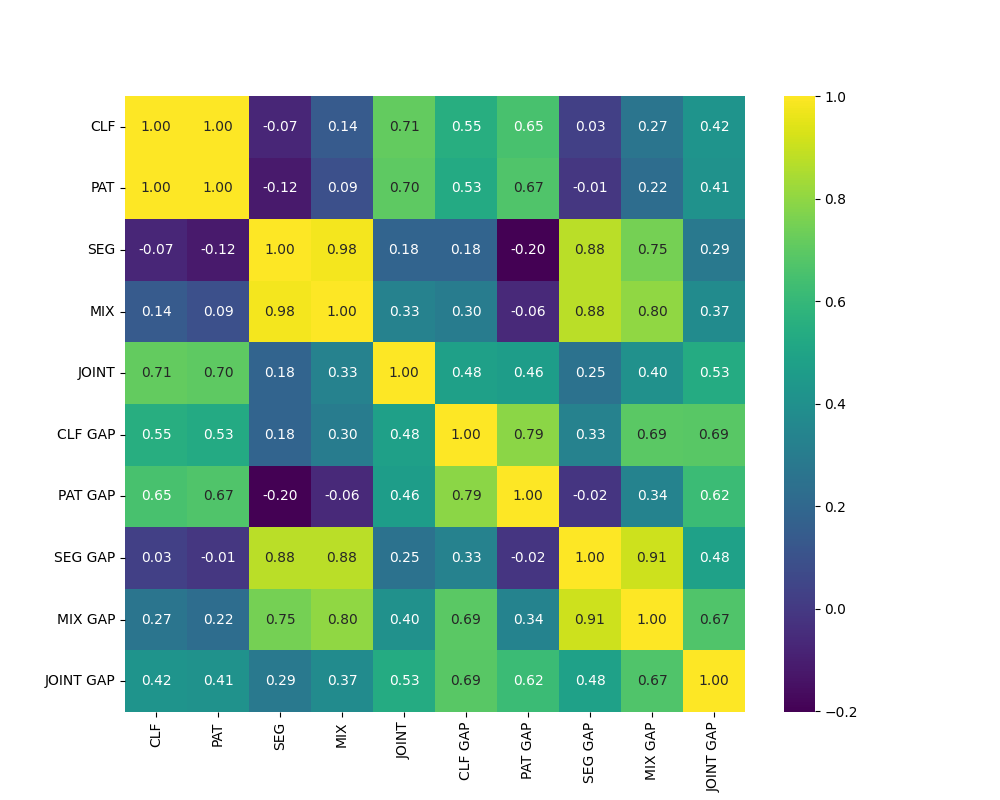}
        \caption{}
        \label{fig:probing_similarity_10}
    \end{subfigure}
    
    \begin{subfigure}{\linewidth}
        \centering
        \includegraphics[width=0.7\linewidth]{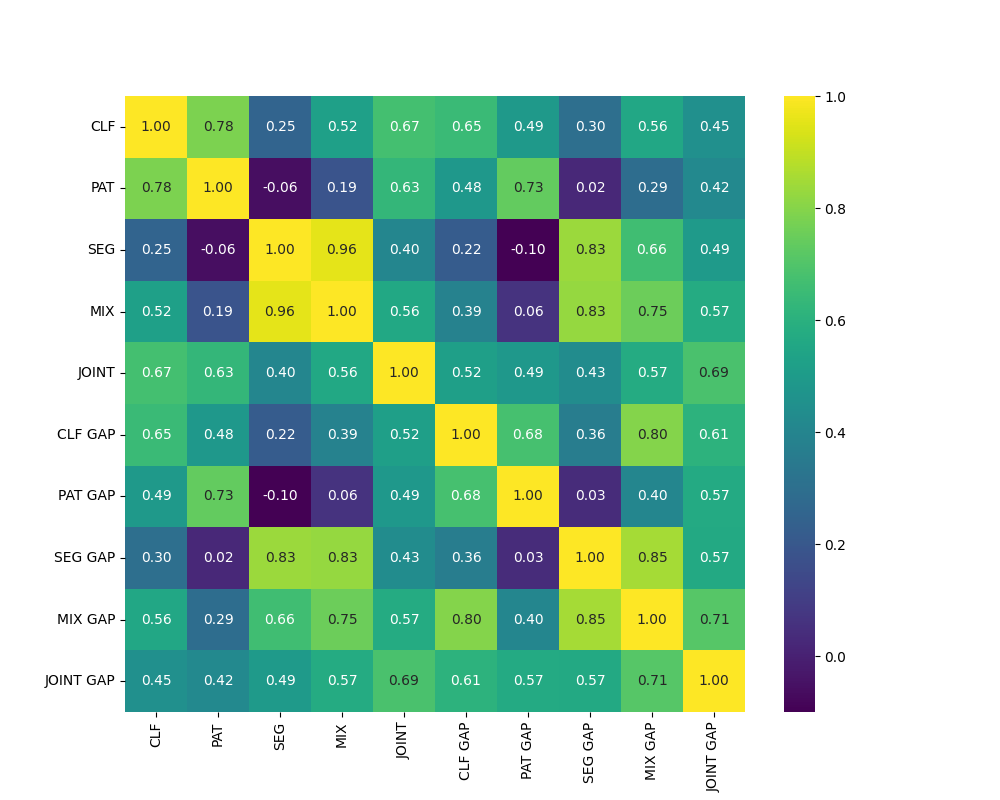}
        \caption{}
        \label{fig:probing_similarity_50}
    \end{subfigure}

    \caption{Cosine similarity matrix showing the average CAV similarity between different probing methods for $148$ tested concepts. Specifically, when (\protect\subref{fig:probing_similarity_10}) probes are trained with only $10$ data points, Classifier- and Pattern-CAVs are approximately identical, while when (\protect\subref{fig:probing_similarity_50}) probes are trained with $50$ data points, the methods learn different representations.}
    \label{fig:probing_similarity}
\end{figure}

\section{Additional feature visualization results}\label{ap:feature_visualization}
\begin{figure}[H]
    \centering
    \includegraphics[width=0.9\linewidth]{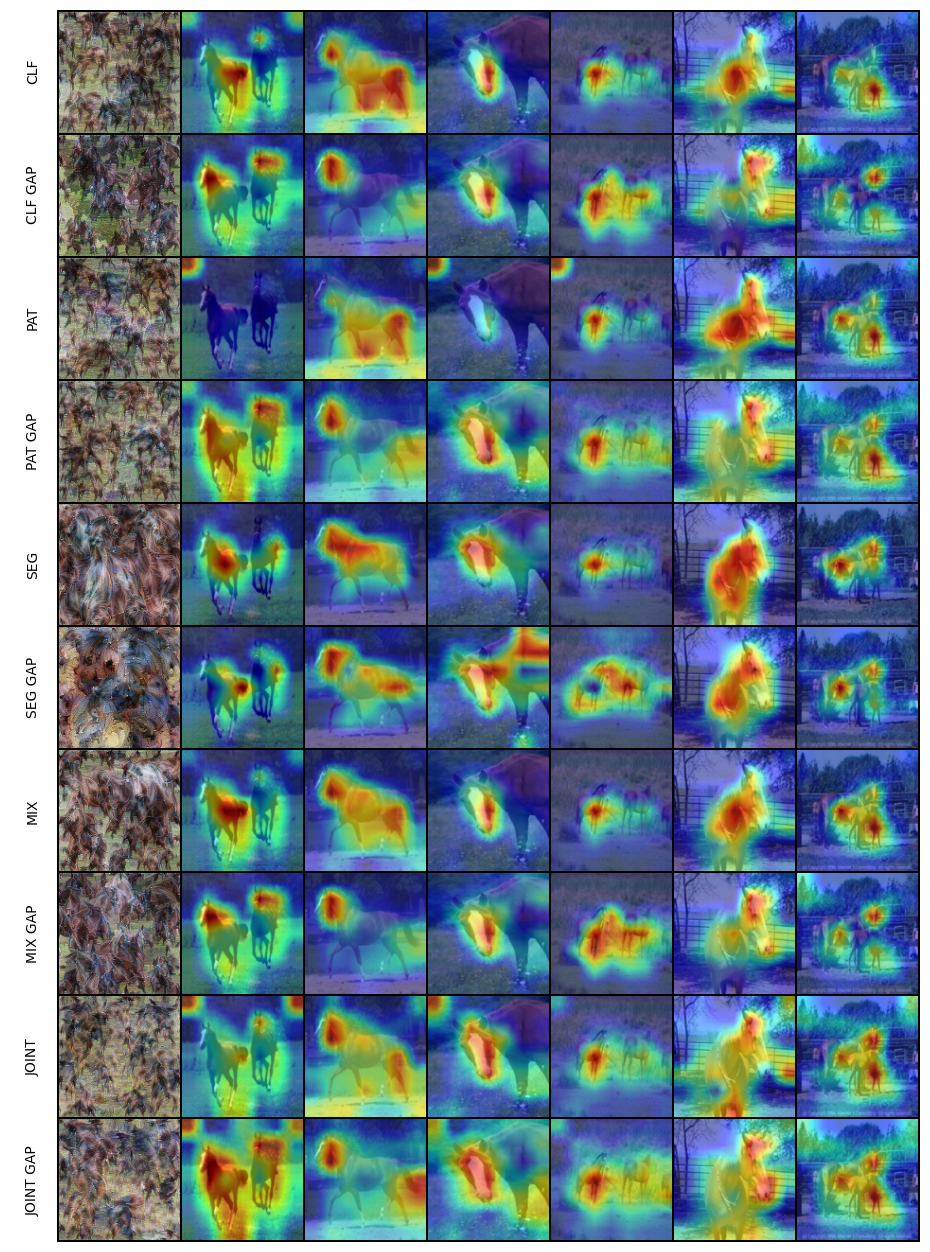}
    \caption{Feature visualization of CAVs obtained by different probing methods, with the first column being activation maximization and the remainder are CLMs for the concept \texttt{horse}.}
    \label{fig:viz_horse}
\end{figure}

\begin{figure}[H]
    \centering
    \includegraphics[width=0.9\linewidth]{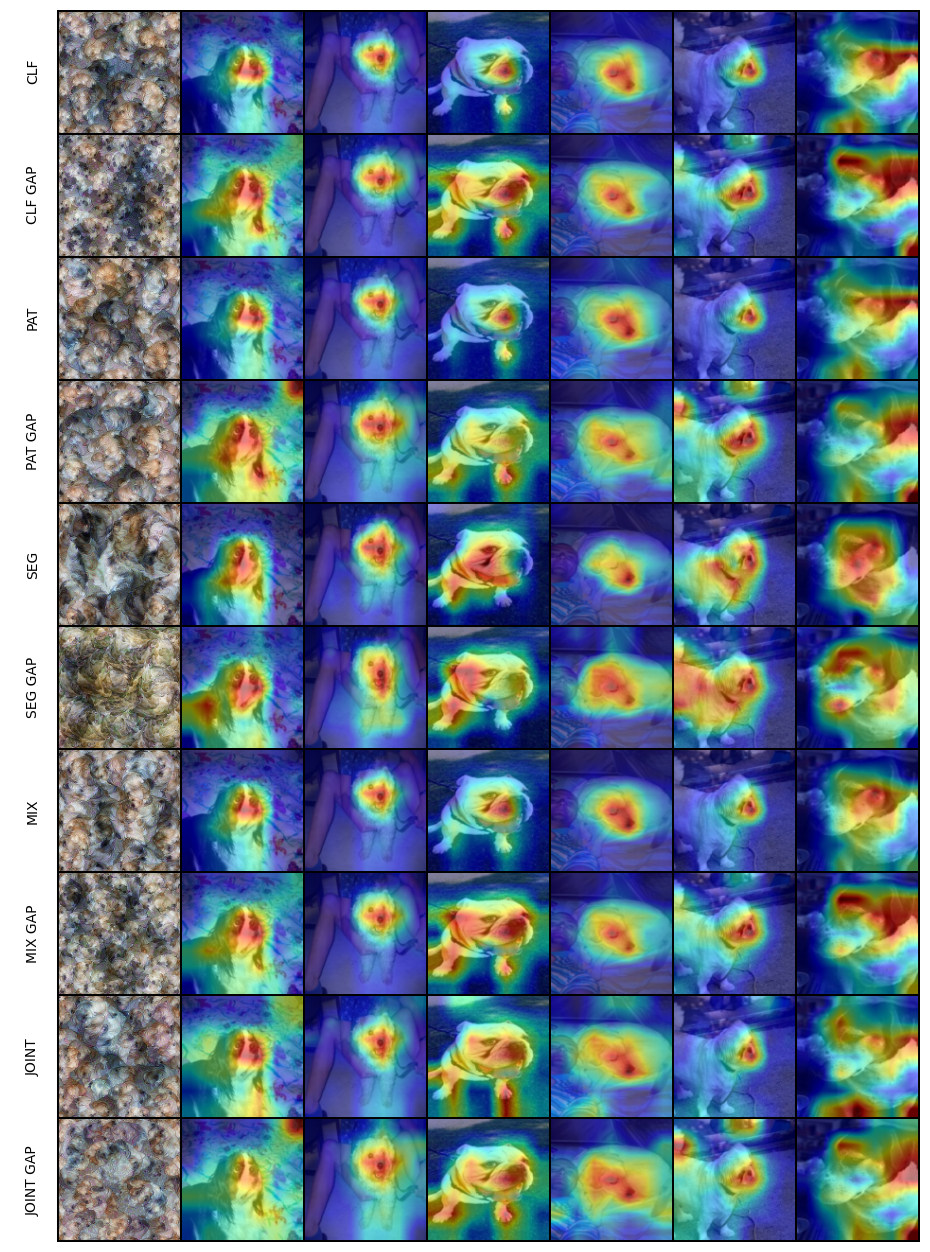}
    \caption{Feature visualization of CAVs obtained by different probing methods, with the first column being activation maximization and the remainder are CLMs for the concept \texttt{dog}.}
    \label{fig:viz_dog}
\end{figure}

\begin{figure}[H]
    \centering
    \includegraphics[width=0.9\linewidth]{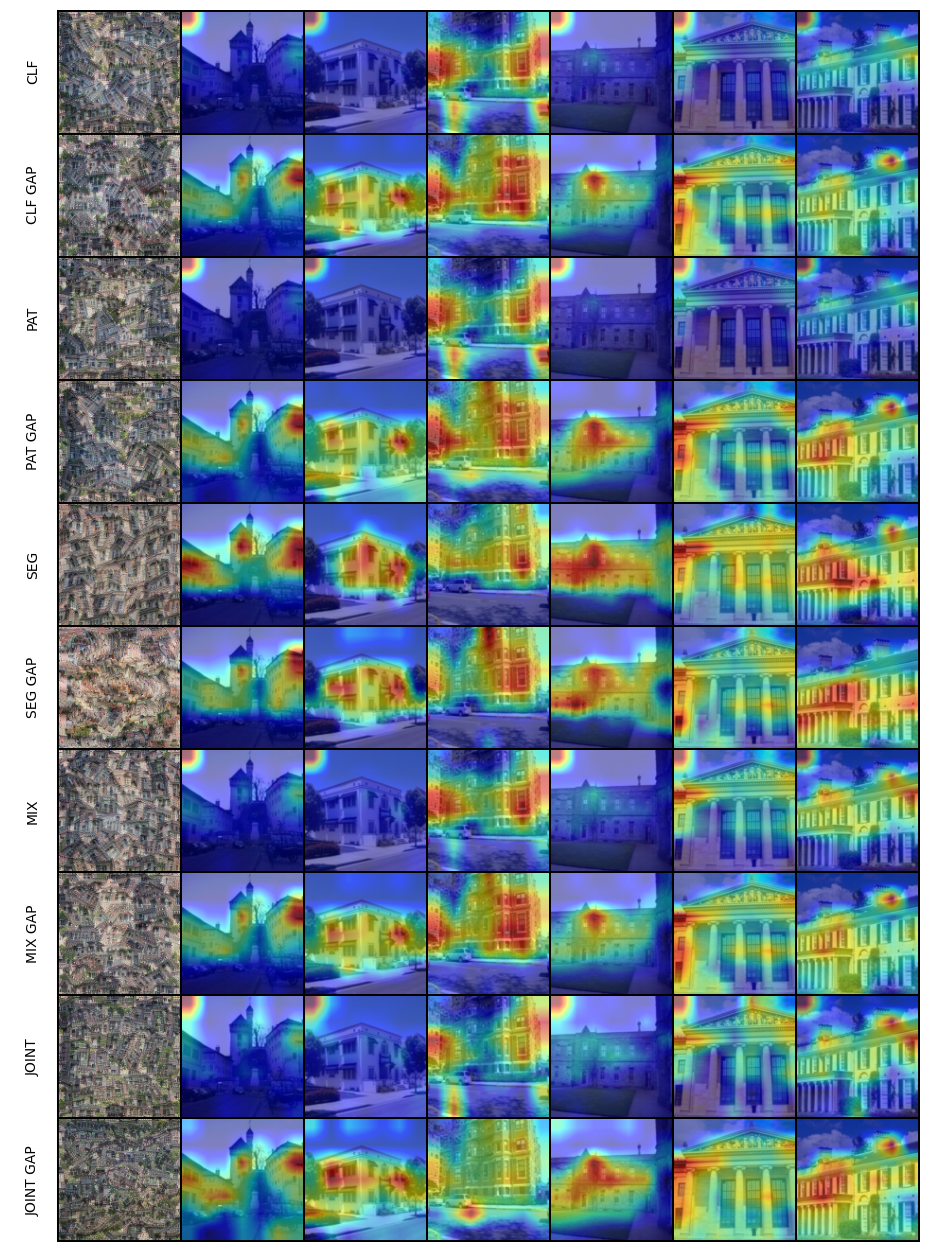}
    \caption{Feature visualization of CAVs obtained by different probing methods, with the first column being activation maximization and the remainder are CLMs for the concept \texttt{building}.}
    \label{fig:viz_building}
\end{figure}

\begin{figure}[H]
    \centering
    \begin{subfigure}{\linewidth}
        \centering
        \includegraphics[width=0.9\linewidth]{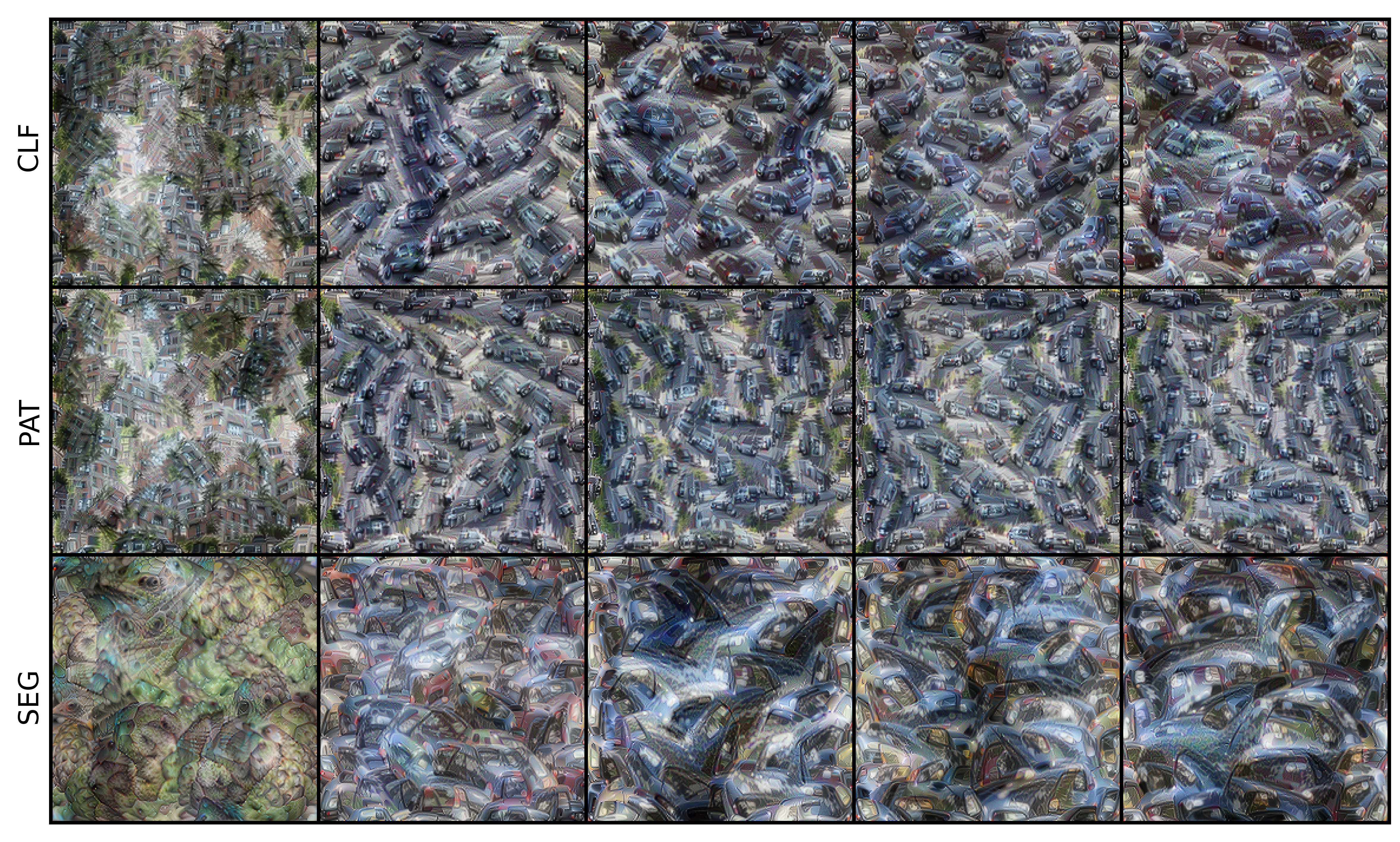}
        \caption{}
        \label{fig:car_scaling}
    \end{subfigure}

    \begin{subfigure}{\linewidth}
        \centering
        \includegraphics[width=0.9\linewidth]{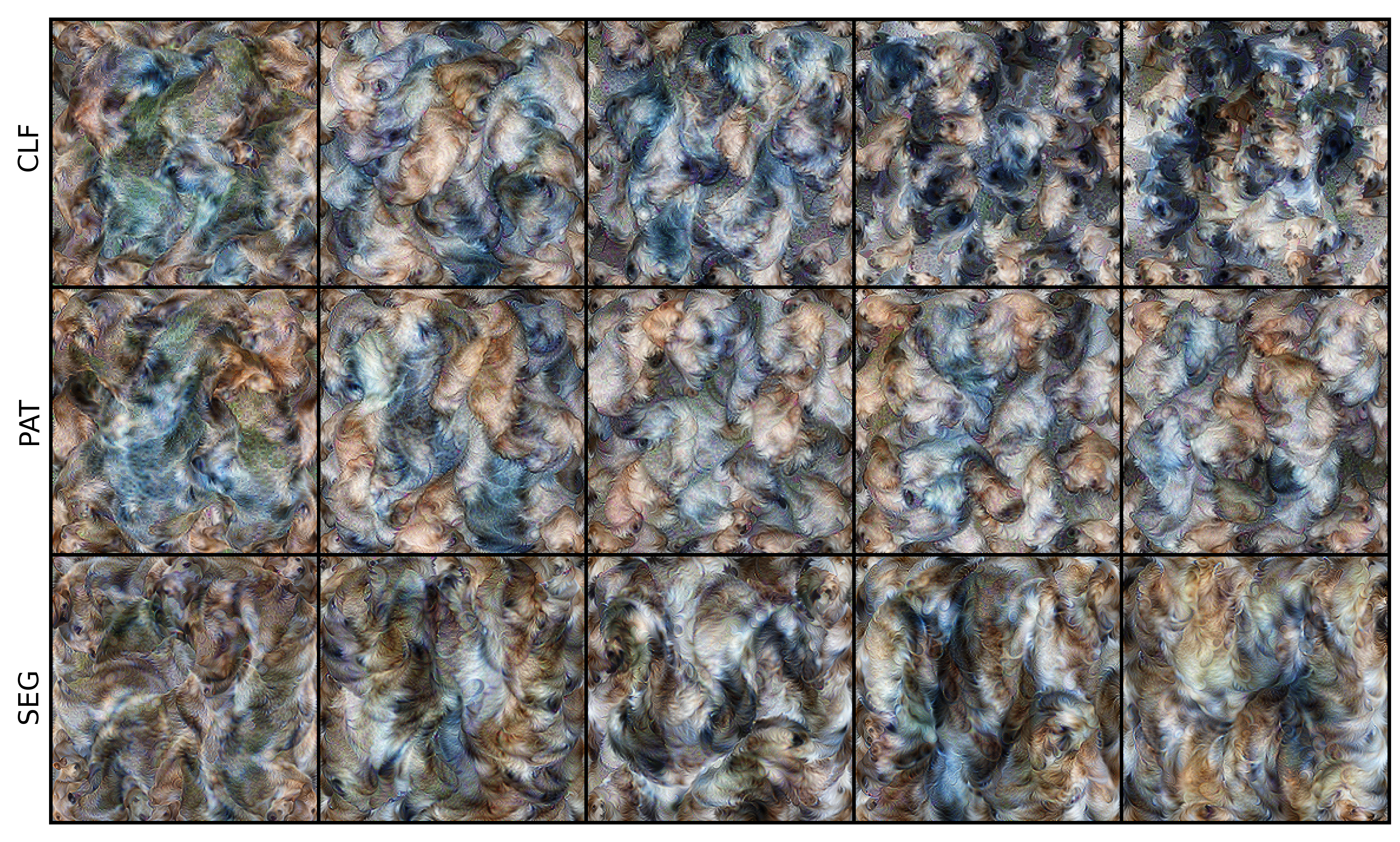}
        \caption{}
        \label{fig:dog_scaling}
    \end{subfigure}

    \caption{Synthetic images generated with activation maximization with different probing methods as rows and different training data sizes $N=\{10, 100, 250, 500, 1000\}$ as columns from left to right. In (\protect\subref{fig:car_scaling}), the probes are trained on the concept \texttt{car} while CAVs in (\protect\subref{fig:dog_scaling}) are trained on the concept \texttt{dog}. Pattern- and Classifier-CAVs show similar features for small training samples, but as data increases, Classifier-CAVs tend to learn more discriminatory features, such as eyes or snouts for the concept \texttt{dog} and fewer spurious correlations for \texttt{car}. In contrast, Pattern-CAVs show minimal change across data sizes.}
    \label{fig:activation_maximization_scaling}
\end{figure}

\section{Additional TCAV results}\label{ap:tcav}
\begin{figure}[H]
    \centering
    \begin{subfigure}{\linewidth}
        \centering
        \includegraphics[width=1.0\linewidth]{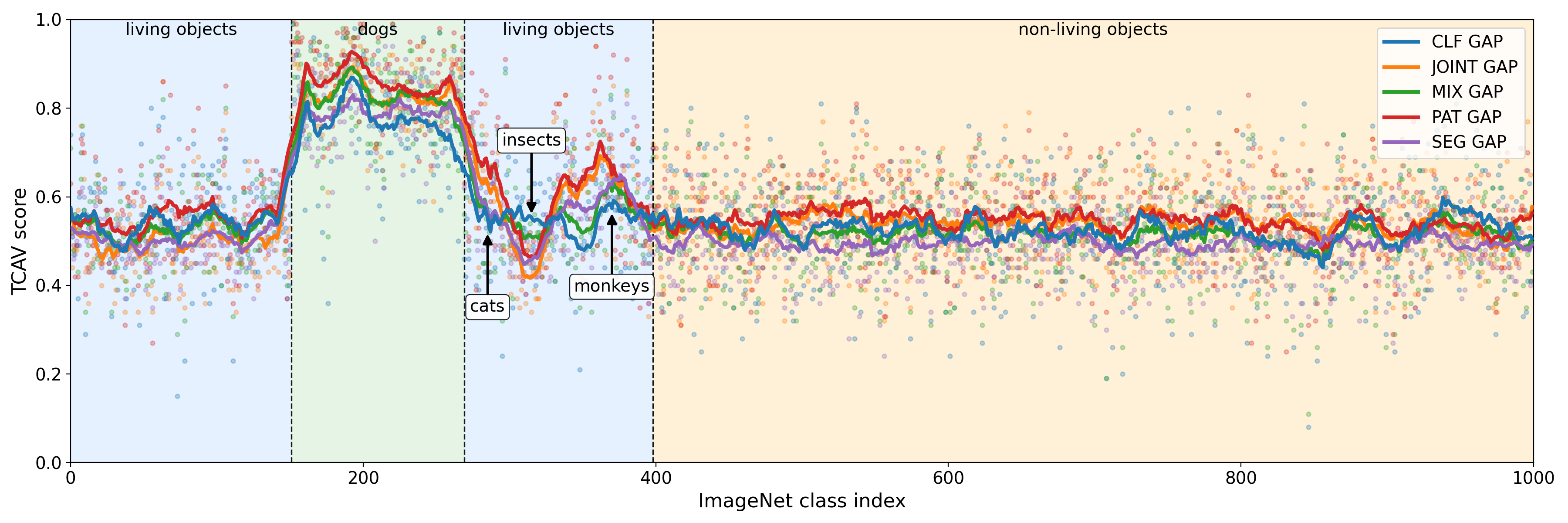}
        \caption{}
        \label{fig:tcav_gap}
    \end{subfigure}

    \begin{subfigure}{\linewidth}
        \centering
        \includegraphics[width=1.0\linewidth]{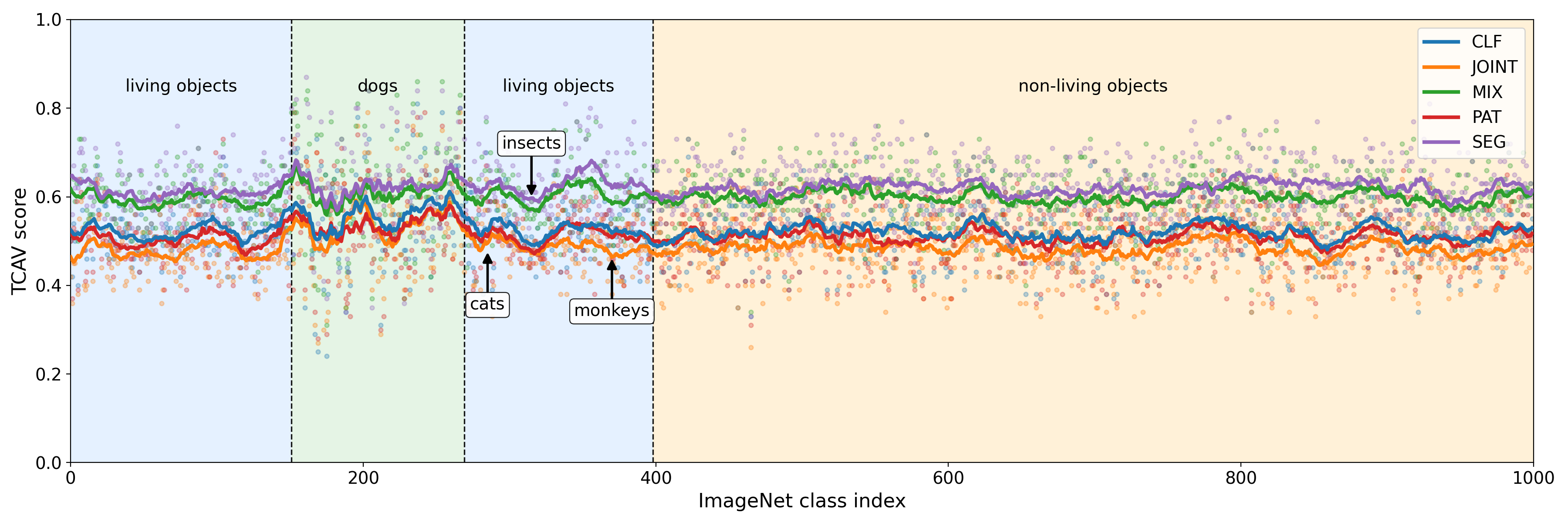}
        \caption{}
        \label{fig:tcav_conv1}
    \end{subfigure}
    
    \begin{subfigure}{\linewidth}
        \centering
        \includegraphics[width=1.0\linewidth]{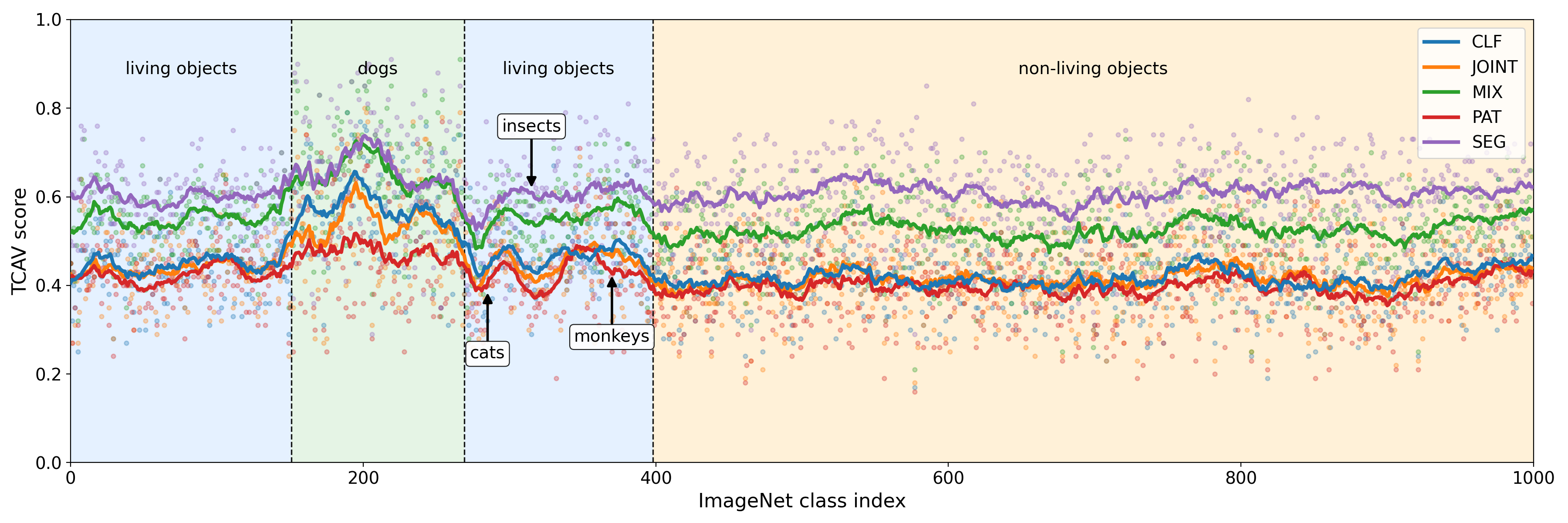}
        \caption{}
        \label{fig:tcav_conv1_500}
    \end{subfigure}

    \caption{$\text{TCAV}$ scores for all ImageNet classes and the concept \texttt{dog} for (\protect\subref{fig:tcav_gap}) translation-invariant probes, (\protect\subref{fig:tcav_conv1}) activations in \texttt{resnet50.layer4[0].conv1} with $50$ training samples, and (\protect\subref{fig:tcav_conv1_500}) activations in the same layer with $500$ training samples.}
    \label{fig:additional_tcav}
\end{figure}

\end{document}

%% file: metrics_table.tex

\begin{table}[htbp]
    \centering
    \renewcommand{\arraystretch}{1.5}
    \caption{Concept alignment metrics for CAVs trained with $N=50$ samples. The experiment is repeated $5$ times, and the mean and standard deviation are reported. The best methods per metric is highlighted in bold.}
    \label{tab:alignment_metrics}
    \resizebox{\textwidth}{!}{
    \begin{tabular}{p{1.5cm}cccccccc}
    \toprule
    \large\textbf{Method} & \large\textbf{Accuracy} & \large\textbf{Hard Accuracy} & \large\textbf{Segmentation} & \large\textbf{Similarity} & \large\textbf{Flip} & \large\textbf{Noise} & \large\textbf{Grayscale} & \large\textbf{Background} \\
    \midrule
    CLF & $0.829 \pm 0.006$ & $0.551 \pm 0.004$ & $0.161 \pm 0.007$ & $0.086 \pm 0.008$ & $0.989 \pm 0.001$ & $0.978 \pm 0.001$ & $0.982 \pm 0.001$ & $0.938 \pm 0.003$ \\
    JOINT & $0.815 \pm 0.007$ & $0.544 \pm 0.005$ & $0.147 \pm 0.007$ & $\mathbf{0.057 \pm 0.003}$ & $0.994 \pm 0.000$ & $0.985 \pm 0.001$ & $0.988 \pm 0.001$ & $0.961 \pm 0.001$ \\
    MIX & $0.831 \pm 0.006$ & $0.571 \pm 0.008$ & $0.193 \pm 0.007$ & $0.177 \pm 0.006$ & $0.992 \pm 0.001$ & $0.978 \pm 0.001$ & $0.985 \pm 0.001$ & $0.957 \pm 0.002$ \\
    PAT & $0.804 \pm 0.010$ & $0.535 \pm 0.006$ & $0.143 \pm 0.005$ & $0.206 \pm 0.010$ & $0.983 \pm 0.001$ & $0.963 \pm 0.002$ & $0.975 \pm 0.001$ & $0.908 \pm 0.003$ \\
    SEG & $0.633 \pm 0.009$ & $\mathbf{0.590 \pm 0.007}$ & $\mathbf{0.231 \pm 0.006}$ & $0.336 \pm 0.012$ & $0.996 \pm 0.000$ & $0.986 \pm 0.000$ & $0.991 \pm 0.000$ & $0.990 \pm 0.000$ \\
    \midrule
    \multicolumn{9}{c}{\text{Translation-invariant methods}} \\
    \midrule
    CLF & $\mathbf{0.848 \pm 0.005}$ & $0.571 \pm 0.006$ & $0.186 \pm 0.009$ & $0.073 \pm 0.011$ & $0.995 \pm 0.000$ & $0.980 \pm 0.002$ & $0.980 \pm 0.003$ & $0.957 \pm 0.002$ \\
    JOINT & $0.821 \pm 0.010$ & $0.547 \pm 0.006$ & $0.152 \pm 0.007$ & $0.150 \pm 0.011$ & $0.996 \pm 0.000$ & $0.980 \pm 0.001$ & $0.982 \pm 0.002$ & $0.956 \pm 0.001$ \\
    MIX & $\mathbf{0.848 \pm 0.005}$ & $0.586 \pm 0.008$ & $0.205 \pm 0.007$ & $\mathbf{0.055 \pm 0.006}$ & $0.996 \pm 0.000$ & $0.982 \pm 0.002$ & $0.984 \pm 0.003$ & $0.968 \pm 0.001$ \\
    PAT & $0.823 \pm 0.012$ & $0.539 \pm 0.006$ & $0.155 \pm 0.007$ & $0.362 \pm 0.023$ & $0.993 \pm 0.000$ & $0.964 \pm 0.003$ & $0.967 \pm 0.003$ & $0.913 \pm 0.002$ \\
    SEG & $0.728 \pm 0.014$ & $\mathbf{0.607 \pm 0.011}$ & $\mathbf{0.225 \pm 0.005}$ & $\mathbf{0.055 \pm 0.003}$ & $\mathbf{0.998 \pm 0.000}$ & $\mathbf{0.991 \pm 0.001}$ & $\mathbf{0.993 \pm 0.001}$ & $\mathbf{0.993 \pm 0.000}$ \\
    \bottomrule
    \end{tabular}}
\end{table}